\documentclass[letterpaper, 10 pt, conference]{ieeeconf}  

\IEEEoverridecommandlockouts                              

\usepackage{amsmath} 
\usepackage{amssymb}  
\usepackage{xcolor}
\usepackage{algorithm}
\usepackage[noend]{algorithmic}
\usepackage{xspace}
\usepackage[font=scriptsize,labelfont=bf]{caption}
\usepackage{subcaption}
\usepackage{graphicx}
\usepackage{times}
\usepackage{soul}
\usepackage{url}
\usepackage{subcaption}
\usepackage{booktabs}
\usepackage{tabularx}
\usepackage{pgf}
\usepackage{array}

\newcommand{\vanillaMCTS}{{{vanilla-HB-MCTS}}\xspace}
\newcommand{\improvedMCTS}{{{HB-MCP}}\xspace}
\newcommand{\daBSP}{{{DA-BSP}}\xspace}
\newcommand{\PFT}{{{PFT-DPW}}\xspace}

\newcommand{\prob}[1]{\ensuremath{\mathbb{P}({#1})}}

\newtheorem{lemma}{Lemma}

\title{\LARGE \bf
Monte Carlo Planning in Hybrid Belief POMDPs}

\author{Moran Barenboim$^{1}$, Moshe Shienman$^{1}$ and Vadim Indelman$^{2}$
\thanks{$^{1}$Moran Barenboim and Moshe Shienman are with the Technion Autonomous Systems Program (TASP), Technion - Israel Institute of Technology, Haifa 32000,
	Israel, {\tt \{moranbar, smoshe\}@campus.technion.ac.il}}%
\thanks{$^{2}$ Vadim Indelman is with the Department of Aerospace Engineering, Technion - Israel Institute of Technology, Haifa 32000, Israel. {\tt vadim.indelman@technion.ac.il}}%
\thanks{ This work was partially funded  
	by US NSF/US-Israel BSF, 
	and by the Israeli Smart Transportation Research Center (ISTRC).}
}

\begin{document}

\maketitle
\thispagestyle{empty}
\pagestyle{empty}

\begin{abstract}
Real-world problems often require reasoning about hybrid beliefs, over both
discrete and continuous random variables. Yet, such a setting has hardly been
investigated in the context of planning. Moreover, existing online Partially
Observable Markov Decision Processes (POMDPs) solvers do not support hybrid
beliefs directly. In particular, these solvers do not address the added
computational burden due to an increasing number of hypotheses with the planning
horizon, which can grow exponentially. As part of this work, we present a novel
algorithm, Hybrid Belief Monte Carlo Planning (\improvedMCTS) that utilizes the
Monte Carlo Tree Search (MCTS) algorithm to solve a POMDP while maintaining a
hybrid belief. We illustrate how the upper confidence bound (UCB) exploration
bonus can be leveraged to guide the growth of hypotheses trees alongside the
belief trees. We then evaluate our approach in highly aliased simulated
environments where unresolved data association leads to multi-modal belief
hypotheses.
\end{abstract}


\section{Introduction}
Intelligent autonomous agents operating in real-world environments often need to
reason about a hybrid belief containing discrete and continuous random
variables. While the states of the agent and of the environment are commonly
represented by continuous random variables, discrete random variables generally
represent object classes, data association hypotheses or even transition models
(e.g. due to slippage) and observation models. In ambiguous environments, where
different objects or scenes can possibly be perceptually similar or identical,
such discrete variables are particularly important, as wrong assignments can
lead to a complete failure of the agent's task.

In general, all random variables in a hybrid belief are coupled, and the number
of hypotheses, i.e. realizations of discrete variables may be combinatorially
large with the number of ambiguous objects and classes or even develop
exponentially with time given ambiguous data associations. Therefore, without
any pruning or merging heuristic, the size of the considered belief quickly
becomes prohibitively large and the computational complexity of the
corresponding problem becomes impossible to handle. 

The research community has been extensively investigating passive inference
approaches where the considered belief is hybrid. In \cite{Segal14iros} the
authors proposed a message passing algorithm to correctly identify loop closures
by optimizing a hybrid factor graph \cite{Kschischang01it}. A convex relaxation
approach over a discrete-continuous graphical model was presented in
\cite{Lajoie19ral} to capture perceptual aliasing and find the maximal subset of
internally coherent measurements, i.e. correct data association. 
\begin{figure}
	\begin{subfigure}{0.48\textwidth}
		\centering
		\includegraphics[scale=1]{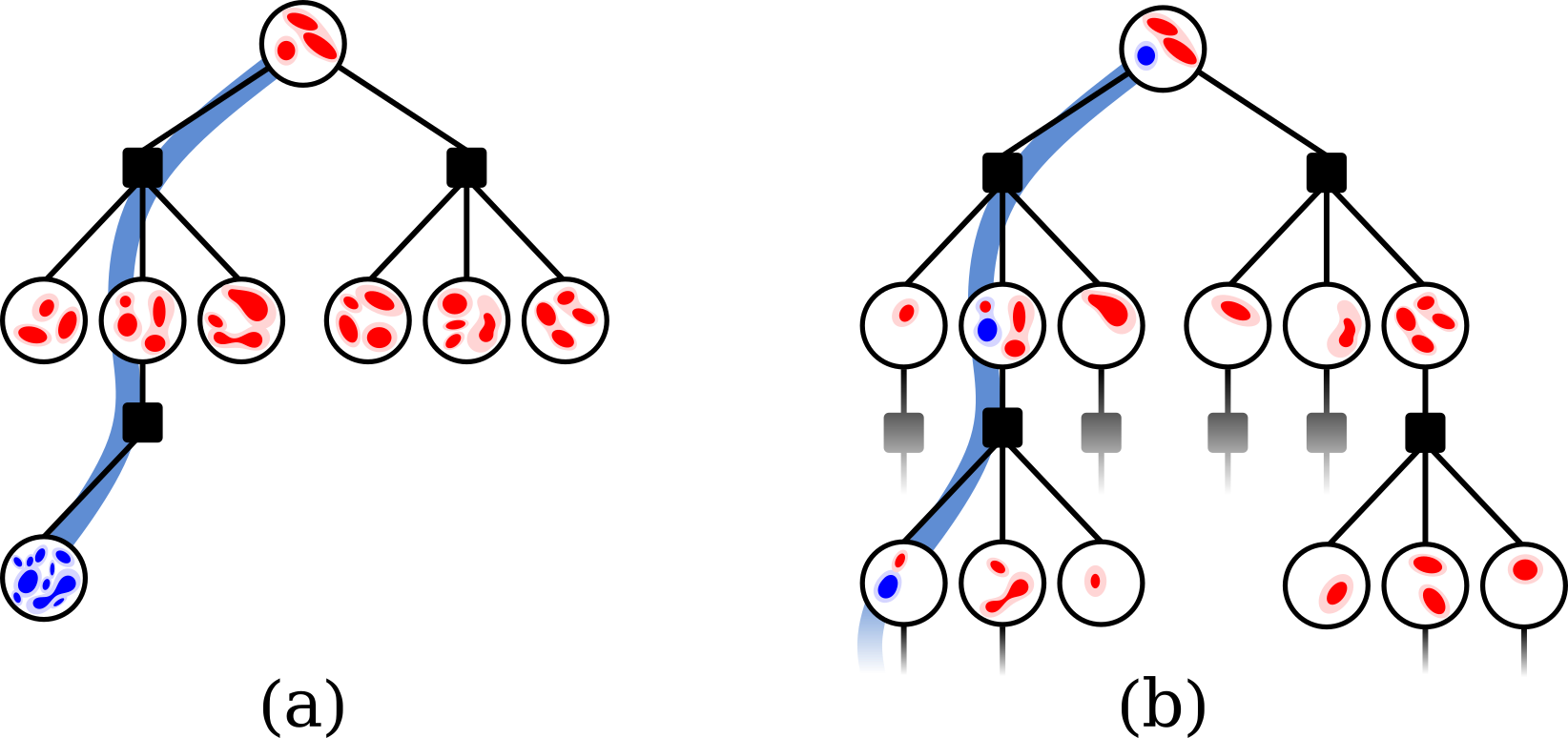}
	\end{subfigure}
	\caption{Nodes in the tree correspond to hybrid beliefs. Inner shapes
	illustrate different continuous distributions, each correspond to a
	different discrete variable. (a) A belief tree which computes a full hybrid
	belief (shown in blue) at each iteration, regardless of the hypotheses
	significance. (b) An adaptive approach (ours) that simultaneously generates
	both hypotheses and nodes in the belief tree sampled according to their
	probabilities.\vspace{-15pt}}
	\label{fig: monte carlo planning}
\end{figure} 

In spite of the significant progress made within the SLAM community, such a
hybrid setting has received scant attention from the planning community. As
such, most off-the-shelf, state-of-the-art POMDP online solvers do not directly
support hybrid beliefs. Specifically, \cite{Silver10nips} introduced POMCP, an
adaptation to Monte-Carlo Tree Search (MCTS) for POMDPs using the UCT algorithm
\cite{Kocsis06ecml} to guide the action selection process. POMCPOW and DESPOT
\cite{Sunberg18icaps, Somani13nips} employ transition and observation models to
efficiently propagate particles from the prior belief, as an efficient
approximation for belief update. However, in the context of hybrid beliefs, the
belief update may not be as efficient, since it would require knowledge of the
hypotheses' probabilities, which are not presumed to be given.

POMDPs can also be converted into belief Markov decision processes (BMDPs) to
utilize MDP solvers. PFT-DPW \cite{Sunberg18icaps} and AI-BSP
\cite{Barenboim22ijcai} are two such solvers, where belief-states replace states
in the original MDP algorithms. However, performing inference with hybrid belief
is hardly efficient due to a large number of hypotheses. For instance, in
ambiguous data association scenarios, the number of hypotheses grows
exponentially with time, making full inference intractable.

Only recently have hybrid beliefs been explicitly considered in planning. In
\cite{Pathak18ijrr}, the authors introduced DA-BSP, which allows reasoning about
future data association hypotheses within a belief space planning framework for
the first time. \cite{Shienman22icra} suggested reducing the computational
complexity of DA-BSP by selecting only a small subset of hypotheses and
providing bounds over the loss in solution quality. \cite{Shienman22icra} was
later extended to a non-myopic setting, in \cite{Shienman22isrr}. The ARAS
framework proposed in \cite{Hsiao20iros} leveraged the graphical model presented
in \cite{Hsiao19icra} to reason about ambiguous data association in future
beliefs using multi-modal factors to model discrete ambiguities. Due to its high
computational burden, these approaches did not aim at closed-loop POMDP
planning, neglecting its mathematical soundness.

In this paper we propose an approach to alleviate the computational complexity
of planning with hybrid beliefs under the POMDP formulation. We show that
previous algorithms result in biased estimators of the reward and value
function, and suggest a different way for controlling the number of hypotheses
to a manageable size. Utilizing sequential importance resampling (SIR) for
hypothesis selection, we suggest an algorithm that results in an unbiased
estimator and efficient belief tree construction. We show that the algorithm
supports both state-dependent and belief-dependent rewards. We proceed with a
contribution to inference in the setting of ambiguous data association, by
introducing a natural way to incorporate negative information within Bayesian
inference, and demonstrate how the hypotheses weights should be updated. Last,
we demonstrate our approach on simulative environments to corroborate our
findings. 

Our contributions in this paper are as follows: (a) We introduce a novel
algorithm that performs Monte-Carlo planning to solve a POMDP when the
considered belief is hybrid. (b) We show that our algorithm, \improvedMCTS,
leads to an unbiased utility estimate, in contrast to existing hybrid belief
algorithms. (c) We introduce negative information to hybrid belief inference.
(d) We demonstrate the effectiveness of our algorithm in extremely aliased
simulated environments where unresolved data association leads to multi-modal
belief hypotheses. This paper is accompanied by supplementary material
\cite{Barenboim23ral_supplementary} that provides proofs and further
implementation and experimental details.

\section{Preliminaries}
In this section we formally define a POMDP and a general hybrid belief, which
will be used in the following sections.

\subsection{Partially Observable Markov Decision Process}
A discrete-time POMDP can be formally defined as a tuple $\left( \mathcal{X},
\mathcal{A}, \mathcal{Z}, T, O, \mathcal{R} \right)$, where $\mathcal{X},
\mathcal{A}$ and $\mathcal{Z}$ denote the state, action and observation spaces
respectively; $T \left( x,a,x' \right) \triangleq \prob{x' | x, a}$ is the
transition density function which expresses the probability to move from state
$x \in \mathcal{X}$ to state $x' \in \mathcal{X}$ by taking action $a \in
\mathcal{A}$; $O \left( x, z \right) \triangleq \prob{z | x}$ is the observation
density function which expresses the probability to receive an observation $z
\in \mathcal{Z}$ from state $x \in \mathcal{X}$; and $\mathcal{R}$ is a user
defined reward function.

As observations provide only partial information about the state, the true state
of the agent is unknown. Therefore, the agent maintains a probability
distribution function over the state space, also known as a belief. At each time
step $t$ the belief update is performed according to Bayes rule, using the
transition and observation models, given the performed action $a_{t-1}$ and the
received observation $z_t$ as $b_t \left( x' \right) = \eta
\int  
\prob{z_t | x' } \prob{x' | x, a_{t-1} } b_{t-1} \left( x \right)  dx$, where $\eta$ is a normalization constant.

Given a posterior belief $b_t$, a policy function $a_t = \pi(b_t)$ determines an
action to be taken at time step $t$. For a finite horizon $\mathcal{T}$ the
value function for a policy $\pi$ is defined as the expected cumulative reward
received by executing $\pi$,
\begin{equation} \label{eq: value function}
	V^{\pi }( b_{t}) =\mathcal{R}( b_{t } ,\pi ( b_{t })) + \underset{z_{t+1:\mathcal{T}}}{\mathbb{E}} \left[ \sum _{\tau =t+1}^{\mathcal{T}}\mathcal{R}( b_{\tau } ,\pi ( b_{\tau }))\right].
\end{equation}
Similarly, an action-value function,
\begin{equation} \label{eq: Q function}
	Q^{\pi }( b_{t}, a_t) = \mathcal{R}( b_t , a_t) + \underset{z_{t+1}}{\mathbb{E}} \left[V^{\pi }( b_{t+1}) \right],
\end{equation}
is defined by executing action $a_t$ and then following the policy $\pi$ for a
finite horizon $\mathcal{T}$. At each planning session, the agent solves a POMDP
by searching for the optimal policy $\pi^*$ that maximizes \eqref{eq: value
function}. 

\subsection{Hybrid Belief}
A hybrid belief is defined over both continuous and discrete random variables.
The continuous random variables can represent the state of the agent and
(possibly also) of the environment, as common in SLAM framework. The discrete
random variables can represent, e.g., object classes and/or data association
hypotheses. Nevertheless, the following definition is general and not restricted
to these examples. 

We formally define the hybrid belief at each time $t$ as
\begin{equation} \label{eq: hybrid belief}
    b_t \triangleq \mathbb{P}(X_t,\beta_{0:t}\mid H_t) = \underbrace{\mathbb{P}(X_t \mid \beta_{0:t}, H_t)}_{b[X_t]_{\beta_{0:t}}} \underbrace{\mathbb{P}(\beta_{0:t}\mid H_t)}_{b[\beta_{0:t}] \equiv \omega_{t}},
\end{equation}
where $X_t\triangleq \{ x_0,..,x_t\}$, $\beta_{0:t}$ denote the discrete random
variables and $H_t \triangleq \{z_{1:t}, a_{0:t-1}\}$ represents all past
actions and observations. $b[X_t]_{\beta_{0:t}}$ is the conditional belief over
continuous variables. $\omega_{t}$ is the marginal belief over discrete
variables which can be considered as the hypothesis weight. We define
$H_{t+1}^{-} \triangleq H_t \cup \lbrace a_t \rbrace$ and $b_{t+1}^{-}
\triangleq \mathbb{P} \left( X_{t+1}, \beta_{0:t+1} | H_{t+1}^{-} \right)$ for
notational convenience.

The marginal belief $\omega_{t}$ is updated for each realization of discrete
random variables according to
\begin{align} \label{eq: recursive weight update}
    &\omega^{i,j}_t 
	\!\!= \!\! \frac{ \overbrace{\mathbb{P}(z_t \mid \beta_{0:t}^{i,j}, H_t^-) \mathbb{P}(\beta_t^i \mid \beta_{0:t-1}^j, H_t^-)}^{\zeta_t^{i|j}} \overbrace{\mathbb{P}(\beta_{0:t-1}^j\mid H_t^-)}^{\omega_{t-1}^j}}{\eta},
\end{align} \normalsize
which is obtained by Bayes rule followed by chain rule on $\omega_{t}$. The
un-normalized weight can be expressed recursively as $\tilde{\omega}^{i,j}_t =
\zeta_t^{i\mid j} \omega^j_{t-1}$. We denote $\beta_t^{i}$ and
$\beta_{0:t-1}^{j}$ as the realization of $\beta_t$ and $\beta_{0:t-1}$
respectively and $\beta_{0:t}^{i,j}$ denotes the realization of the joint
variables $\beta_{0:t}$. $\zeta_t^{i\mid j}$ denotes the conditional dependence
of $\zeta_t$ on $\beta_t^i$ given $\beta_{0:t-1}^j$. This notation proceeds
similarly for any random variable. The conditional belief
$b[X_t]_{\beta_{0:t}}$ is updated for each realization of discrete random
variables as
\begin{equation}\label{eq: conBeliefUp}
	b[X_t]_{\beta_{0:t}}^{i,j} = \psi(b[X_t]^j_{\beta_{0:t-1}}, a_{t-1}, z_t),
	\vspace{-5pt}
\end{equation}
where $\psi(.)$ represents the Bayesian inference method.  

Generally, when planning with hybrid beliefs the agent constructs both a belief
tree and multiple hypotheses trees. Each hypotheses tree represent the posterior
hypotheses given a history. Since every node of the planning tree (i.e. belief
tree) corresponds to a hypotheses tree, the computational complexity of the
corresponding POMDP becomes a significant burden. In the following section we
present a novel algorithm that circumvent this difficulty via Monte-Carlo
sampling.

\subsection{Ambiguous data association}
One use case for hybrid beliefs is ambiguous data associations (DA), which
introduces a significant computational burden in both inference and even more so
in planning. In inference, the number of possible hypotheses grows exponentially
with the number of steps the agent took, i.e. the length of the history array.
In planning, the difficulty further increases due to the exponential number of
histories considered by the agent.

To address ambiguous DA, \eqref{eq: recursive weight
update} can be adapted to, 
\begin{equation} \label{eq: DA weight update}
    \omega_t^{i,j} = \tilde{\zeta} _{t}^{j\mid i} \omega _{t-1}^{i},\vspace{-10pt}
\end{equation}
\vspace{-10pt}where,
\begin{align}
    \tilde{\zeta} _{t}^{j\mid i} &= \frac{\zeta _{t}^{j\mid i}}{\sum_i \zeta _{t}^{j\mid i} \omega_{t-1}^i }, \label{eq: recursive DA weight update} \\ 
    \zeta _{t}^{j\mid i} \!&=\! \int _{X_{t}}\mathbb{P}( z_{t} \mid X_{t},\beta_t^i )\mathbb{P}( \beta _{t}^i \mid X_{t})\mathbb{P}( X_t \mid \beta _{0:t-1}^j ,H_{t}^-). \notag
\end{align}
The latter is obtained by marginalizing $\zeta_t^{i\mid j}$ over the states, and
adhering to the Markov assumption of the observation and association
($\mathbb{P}(\beta_t\mid X_t)$) models, as suggested in \cite{Pathak18ijrr}.
Our hybrid belief planner is tested in section \ref{sec:experiments} using ambiguous DA as a challenging use
case.

\section{POMDP Planning with hybrid beliefs}
In section \ref{sec:vanillaMCTS} we start with a brief overview of how MCTS
can be utilized to solve POMDPs with hybrid beliefs and its drawbacks. Then,
in section \ref{sec:improvedMCTS} we
present a novel approach to utilize the UCT exploration bonus to build an
asymmetric hypotheses tree, which leads to better use of the computational
resources by focusing on the most promising hypotheses according to the UCT
bonus.

\subsection{vanilla Hybrid-Belief MCTS}   \label{sec:vanillaMCTS} For
completeness, we first present a \vanillaMCTS algorithm. Although the exact
algorithm does not seem to exist in the literature, this is the ad-hoc way to
interleave hybrid beliefs with state-of-the-art POMDP solvers. \vanillaMCTS, can
be seen as an adaptation of the state-dependent MCTS \cite{Silver10nips}
algorithm to a (hybrid-)belief \eqref{eq: hybrid belief}, by augmenting the
belief to a belief-state. A similar approach was also taken by PFT-DPW
\cite{Sunberg18icaps}, which utilized particle filters to approximate a
posterior belief, over continuous variables. However, computing a full hybrid
belief is a difficult and sometimes intractable task, even for particle-based
solvers, and is thus prone to approximations.

\textbf{Pruning.} The number of hypotheses at each posterior node in the belief
tree may be prohibitively large. To handle the infeasible number of the
posterior hypotheses, \vanillaMCTS utilizes a pruning mechanism similar to those
suggested in \cite{Pathak18ijrr,Hsiao19icra}. As a result, unlikely hypotheses
are removed from the hypotheses tree.

In \vanillaMCTS, each posterior node holds a fixed number of hypotheses once
expanded, depending on a predefined hyperparameter. Such a method may sometimes
be too harsh, pruning away hypotheses with high probability due to a limited
hypotheses budget, or too loose, keeping highly unlikely hypotheses, thus
wasting valuable computational time. Other approaches may also be applicable,
such as fixing a probability threshold value, under which all hypotheses are
pruned. However, the latter has its own deficiencies, such as hypothesis
depletion. For completeness, we describe \vanillaMCTS implementation details in
the supplementary  \cite{Barenboim23ral_supplementary}.

\subsection{Hybrid Belief Monte-Carlo Planning}\label{sec:improvedMCTS}

In contrast to \vanillaMCTS, in \improvedMCTS, we do not use any pruning
heuristic for two reasons: (1) this requires knowledge, or an insight, as to how
many hypotheses would be sufficient for the specific POMDP; (2) Each posterior
node in the belief tree maintains hypotheses based on a hyperparameter,
regardless of how relevant this node may be for decision-making. 

Conversely, we suggest an adaptive algorithm that focuses computational
resources in proportion to their relevance in the belief tree, which circumvent
the difficulty in full belief update. \improvedMCTS is recursively invoked with
a single sampled hypothesis. Every such single hypothesis may evolve into
multiple hypotheses. \improvedMCTS algorithm computes only the posterior weights
(i.e. probability values) that are conditioned on that single hypothesis,
followed by a random weight sample based on their categorical distribution.
Then, only the hypothesis associated with the sampled weight is updated. This is
in contrast to the full posterior update done in \vanillaMCTS.

Additionally, to support belief-dependent rewards, the reward value is estimated
based on state samples received across multiple visits to the belief node, i.e.,
state samples from multiple hypotheses. We describe the algorithm details in
section \ref{sec:algorithms}.

\improvedMCTS holds some desirable properties compared to the full belief update
and pruning approaches. First, at each iteration of \improvedMCTS, a maximum of
$\mathcal{T}$ posterior hypotheses are computed, and a small subset of the
weights. This is in contrast to the full posterior update, that would require
the entire (or pruned-)set of the current posteriors, and compute all the
posterior hypotheses of the next time-step, which is highly resource expensive
for every iteration. Second, \improvedMCTS explores both the planning tree and
the hypotheses trees by focusing its computational effort on the interesting
parts, utilizing UCB to guide the search; this property is inspired by MCTS
which builds the planning tree by focusing on the optimistic parts of the tree.
In section \ref{sec:obj}, we show that this approach results in an unbiased
estimator for the true value function.

\section{Implementation details}\label{sec:algorithms}
In this section we describe the implementation details of our approach,
\improvedMCTS, as discussed in section \ref{sec:improvedMCTS}. 

\begin{algorithm}[t] 
	{\scriptsize
    \caption{\improvedMCTS} 
    \textbf{Procedure}:\textsc{Simulate}($b_t^j,h,d$) \label{alg:MC-HBT}
    \begin{algorithmic}[1] 
        \IF{d = 0}
        \RETURN 0
        \ENDIF
        \STATE $a \xleftarrow{} \underset{\bar{a}}{\arg \max} \ Q(h\bar{a}) +
        c\sqrt{\frac{log(N(h))}{N(h\bar{a})}}$
        \STATE $B(h) \leftarrow $\textsc{GetSamples}($b_t^j, B(h), N(h)$)
        \label{alg: GetSamples}
        \STATE $r \xleftarrow{}$ \textsc{Reward}$(B(h), a)$ 
        \STATE $r \xleftarrow{} r + N(h)(r-r_{prev})$ \label{alg: rwrd}
        \IF{$|C(ha)| \leq k_oN(ha)^{\alpha_o}$} \label{alg: progressive start}
        \STATE $z \leftarrow$  \textsc{SampleObservation}$(b_t^j, a)$ \label{alg: sample observation}
        \ELSE{}
        \STATE $z \leftarrow$  {Sample uniformly from $C(ha)$} \label{alg: progressive end}
        \ENDIF
        \STATE $\{\omega_{t+1}^{i, j}\}_{i=1}^L \xleftarrow{}$
        \textsc{ComputeWeights}$(b_t^j, a, z)$ \label{alg: Compute weights}
        \STATE $i \xleftarrow{}$ \textsc{SampleCategorical}$(\{\omega_{t+1}^{i,
        j}\}_{i=1}^L)$ \label{alg: sample hypothesis}
        \STATE $b_{t+1}^{i, j} \xleftarrow{} \Psi(b^j_t,a,z,i)$ \textit{//
        Eq.~\eqref{eq: conBeliefUp}} \label{alg: update hypothesis}
        \IF{ $z \notin C(ha)$} \label{alg: recurse start}
        \STATE $C(ha)\cup \{z\}$
        \STATE $R \xleftarrow{} r +$\textsc{Rollout}$(b_{t+1}^{i,j}, d-1)$
        \ELSE{}
        \STATE $R \xleftarrow{}  r +$\textsc{Simulate}$(b_{t+1}^{i,j}, haz,
        d-1)$ \label{alg: recurse end}
        \ENDIF
        \STATE $N(h) \xleftarrow{} N(h) + 1$ \label{alg: update params start}
        \STATE $N(ha) \xleftarrow{} N(ha) + 1$
        \STATE $Q(ha) \xleftarrow{} Q(ha) + \frac{R-Q(ha)}{N(ha)}$ \label{alg: update params end}
        \STATE \textbf{return} $R$
    \end{algorithmic}
}
\end{algorithm}

\improvedMCTS can be described as follows; first, it starts by receiving a
single hypothesis and selecting an immediate action according to UCB exploration
bonus. Then, samples are generated and appended to $B(h)$, which are later used
for reward estimation (lines \ref{alg: GetSamples}- \ref{alg: rwrd}). Lines
\ref{alg: progressive start}-\ref{alg: progressive end} perform observation
progressive widening. Then, the approach for sampling hypotheses is shown in
lines \ref{alg: Compute weights}-\ref{alg: update hypothesis}. Note that the
algorithm directly computes \textit{all} the weights conditioned on the
hypothesis given as input (line \ref{alg: Compute weights}). 
Then, we resample a \textit{single} conditional belief, $b_{t+1}^{i,j}$,
sampled according to the weights (line \ref{alg: sample hypothesis}). We note
that this is not a necessity, and different number of samples can be taken in
those two steps to trade-off efficiency and accuracy. Depending on whether a new
posterior node is sampled or not, lines \ref{alg: recurse start}-\ref{alg:
recurse end} either call for rollout or continues recursively. Last, the
action-value function and the counters are updated. 

To estimate a belief-dependent reward, state samples should correspond to their
likelihood in the full hybrid belief. In \improvedMCTS, hypotheses are generated
iteratively, accumulating hypotheses (or, equivalently, state samples from those
hypotheses), so that at each iteration the reward estimator is improved.
Generally, a belief dependent reward is not a simple average over samples.
However, as in MCTS, \improvedMCTS estimates the action value function, $Q(ha)$,
as an average of all the cumulative returns passed through that node. To support
belief dependent rewards, \improvedMCTS computes a new reward estimate based on
all past samples, and replaces the previous reward estimate with the new one. To
that end, a simple recursive subtraction and addition update is done for every
node encountered along the path of the current iteration, described in line
\ref{alg: rwrd}.

\section{Theoretical analysis} \label{sec:obj}
In this section, we first claim that existing approximations, done in
contemporary state-of-the-art multi-hypotheses planners, such as DA-BSP
\cite{Pathak18ijrr}, ARAS \cite{Hsiao20iros} as well as \vanillaMCTS
(Section \ref{sec:vanillaMCTS}), lead to a biased estimation of the reward value, and
therefore a biased value function. Further, we show that even if the reward
value could be precisely recovered, the resultant value function is generally
biased. Instead, \improvedMCTS performs sequential sampling which converges to
the correct value. Then, we discuss how \improvedMCTS may also support
belief-dependent reward functions and its applicability for value function
estimation.

\subsection{State-dependent rewards} \label{sec: state dept reward}
State-dependent reward functions are defined as the expected reward value over
the belief, i.e., $\mathcal{R}_X\triangleq\mathbb{E}_{X\sim b}[r(X,a)]$.
Generally, state-dependent rewards cannot be computed analytically, thus, they
are approximated using state samples. Since in a hybrid belief the number of
hypotheses may be prohibitively expensive to compute, most existing algorithms
approximate the belief, $\hat{b}$, by performing some heuristic pruning. As a
consequence, the approximate distribution is shifted, and the reward value is
biased even with an infinite number of state samples, 
\begin{lemma} \label{lemma1}
    The estimator $\mathbb{E}_{X\sim \hat{b}}[r(X,a)]$ is 
    biased.
\end{lemma}
\begin{proof} \label{proof:lemma1}
    Assuming the weights of the pruned hypotheses are non-zero, the proof is
    immediate,
    \begin{align}
        &\mathbb{E}_{X\sim b}[r(X,a)] =\int _{X} \sum _{\beta } b( X, \beta) r( X,a) dX   \\
        &\!=\! \int\limits_{X}\sum_{\beta\in A} b( X, \beta)r(X,a)dX\!+\!\!\sum_{\beta\in \neg A} b( X, \beta)r(X,a) dX  \notag \\
        &\neq \eta_A \int\limits_{X}\sum_{\beta\in A} b( X, \beta)r(X,a)dX= \mathbb{E}_{X\sim \hat{b}}[r(X,a)].\notag
    \end{align}
    where $A$ denotes the set of un-pruned hypotheses, and $\eta_A$ is their
    corresponding normalizer after pruning.
\end{proof} 

In contrast, \improvedMCTS samples hypotheses iteratively starting from the root
node; it utilizes sequential importance resampling, which results in an unbiased
estimator for the reward value. At every iteration, the new sampled states from
the current hypothesis are added to the estimator from previous iterations, by
averaging. The process for generating hypotheses can be described as follows;
for any time $t$, a hypothesis is sampled i.i.d from a proposal-prior
distribution, $\beta_0^i \sim \mathbb{Q}(\beta_0 \mid H_0)$. Then, hypotheses are
recursively sampled from a proposal distribution, $\beta_\tau^i \sim
\mathbb{Q}(\beta_\tau \mid \beta_{0:\tau-1})$ up to time $\tau\!=\!t$. We define
$\mathbb{Q}(\beta_0\mid H_0)\!\triangleq \!\mathbb{P}(\beta_0\mid H_0)$, and
$\mathbb{Q}(\beta_\tau \mid \beta_{0:\tau-1})\triangleq \textsc{Uniform}\left[1,
\left|\beta_\tau\right|\right]$. Then, for every time-step $t$, the
corresponding importance weight is,
\begin{align}
    \lambda_t^{i,j} \!&=\! \frac{\mathbb{P}(\beta_{0:t}^{i,j}\mid H_t)}{\mathbb{Q}(\beta_{0:t}^{i,j}\mid H_0)}
    \!=\! \frac{\eta_t \zeta_t^{i\mid j} \mathbb{P}(\beta_{0:t-1}^{j}\mid H_{t-1})}{\mathbb{Q}(\beta_t^i \mid \beta_{0:t-1}^j)\mathbb{Q}(\beta_{0:t-1}^j\mid H_{0})} \\ 
    &\!=\! \frac{\eta_t \zeta_t^{i\mid j} }{1/ | \beta_t^{i\mid j} | } \frac{\mathbb{P}(\beta_{0:t-1}^j\mid H_{t-1})}{\mathbb{Q}(\beta_{0:t-1}^j\mid H_{0})}
    \!=\!  \eta_t \zeta_t^{i\mid j} |\beta_t^{i\mid j}| \lambda_{t-1}^j, \notag
\end{align}
where $\lambda_0^j=1$. As a consequence,
\begin{lemma} \label{lemma2}
    \improvedMCTS state-dependent reward estimator,
    $\hat{\mathcal{R}}_X\triangleq \frac{1}{N}\sum_{i,j=1}^N \lambda_t^{i,j}
    \frac{1}{n_X}\sum_{k=1}^{n_X} r(X_t^{i,j,k},a_t)$, is unbiased.
    
    \begin{proof} If states are sampled i.i.d. for each hypothesis, then the
    expected value of the reward estimator, $\hat{\mathcal{R}}_X$, is,
    \begin{align} \label{eq: reward estimation}
        &\mathbb{E}\left[\hat{\mathcal{R}}_X\right] \triangleq  \mathbb{E}\left[\frac{1}{N}\sum_{i,j=1}^N \lambda_t^{i,j} \frac{1}{n_X}\sum_{k=1}^{n_X} r(X_t^{i,j,k},a_t)\right] \\
        &=\mathbb{E}_{\mathbb{Q}}\left[\frac{1}{N}\sum_{i,j=1}^N \lambda_t^{i,j} \mathbb{E}_{b[X_t]_{\beta_{0:t}}^{i,j}}\left[\frac{1}{n_X}\sum_{k=1}^{n_X} r(X_t^{i,j,k},a_t)\right]\right] \notag\\
        &= \frac{1}{N}\sum_{i,j=1}^N \mathbb{E}_{\mathbb{Q}} \left[ \frac{\mathbb{P}}{\mathbb{Q}} \frac{1}{n_X}\sum_{k=1}^{n_X}\mathbb{E}_{b[X_t]_{\beta_{0:t}}}\left[ r(X_t^{i,j,k},a_t)\right] \right]\notag \\
        &= \mathbb{E}_{\mathbb{P}} \left[\mathbb{E}_{b[X_t]_{\beta_{0:t}}} r(X_t,a_t) \right]  \triangleq \mathcal{R}_X \notag
    \end{align} 
    where $\mathbb{P}\!=\!\mathbb{P}(\beta_{0:t}\mid H_t)$,
    $\mathbb{Q}\!=\!\mathbb{Q}(\beta_{0:t}\mid H_t)$, and $N$ and $n_X$ denote the number
    of samples from $\mathbb{Q}$ and $b[X_t]^{i,j}_{\beta_{0:t}}$ respectively.
\end{proof}
\end{lemma}
As the planning horizon grows, sampling hypotheses uniformly quickly induce
sample degeneracy. That is, the weights of most hypothesis samples become
negligible, while only a few remain significant, which negatively affects the
accuracy of the estimate. To avoid this issue, we perform resampling at every
step, also known as sequential importance resampling (SIR). Before resampling,
each hypothesis weight simply becomes, $\lambda^{i \mid j}_t= \eta_t
\zeta_t^{i\mid j} \left| \beta^{i \mid j}_t \right| $, which is then updated to
$1/N$ after resampling. Note that resampling does not introduce bias to the
estimator \cite{kennedy16book}. To avoid repeated derivations, for the rest of
this sequel we treat mathematical proofs as if hypotheses are directly sampled
from distribution $\mathbb{P}$, even though they are in fact sampled from the
proposal distribution, $\mathbb{Q}$. However, all derivations can be started by
sampling from $\mathbb{Q}$, then follow similar steps of lemma \ref{lemma2}
followed by resampling to arrive at the same result.

In some cases of interest, such as ambiguous DA, the normalizer $\eta_t$ cannot
be easily computed, and so the importance weight, $\lambda_t$, cannot be
computed. A common practice is to use the self-normalized version of the
estimator, i.e. $\tilde{\lambda}^{i \mid j}_t = \tilde{\lambda}^{i \mid j}_{t-1}
\frac{\zeta_t^{i \mid j}}{\sum \zeta_t^{i \mid j}}$, which is no longer unbiased
\cite{kennedy16book}. However, the self-normalizing variation is consistent,
meaning it becomes less biased with more samples and converges in probability
(denoted $\rightarrow^p$) to the theoretical value. This is a direct consequence
of applying the weak law of large numbers on both the nominator and denominator
of the self-normalized estimator,
\begin{align}
    &\hat{\mathcal{R}}^{SN}_X\triangleq \frac{ \sum_{i,j=1}^N  \zeta_t^{i \mid j} \omega^j_{t-1} \frac{1}{n_X}\sum_{k=1}^{n_X} r(X_t^{i,j,k},a_t)}{ \sum_{i,j=1}^N \zeta_t^{i \mid j} \omega^j_{t-1}} \\
    &=\frac{ \frac{1}{N} \sum_{i,j=1}^N  \eta_t \zeta_t^{i \mid j} \omega^j_{t-1} \frac{1}{n_X}\sum_{k=1}^{n_X} r(X_t^{i,j,k},a_t)}{\frac{1}{N} \sum_{i,j=1}^N \eta_t \zeta_t^{i \mid j} \omega^j_{t-1} }\rightarrow^p \frac{\mathcal{R}_X}{1}, \notag
\end{align}
where the denominator converges to the sum of weights,
$\sum_{i,j}\omega_t^{i,j}=1$ and the nominator to the reward value.

\subsection{Belief-dependent rewards}
Contrary to state-dependent rewards, belief dependent rewards are not
necessarily linear in the belief, so averaging over state samples from different
hypotheses does not guarantee convergence to the theoretical reward value.
Moreover, different reward definitions may be functions of not only the
states, but also the weights, the conditional beliefs, or the probability
density values of the complete theoretical belief (such as Shannon's entropy
\cite{Shienman22icra} or differential entropy \cite{Barenboim22ijcai}). To
support the various cases, we split our discussion into the parametric case,
where the reward can be precisely calculated given a set of parametric
conditional beliefs and the corresponding weights, and the nonparametric case,
where the reward is estimated based on state and hypothesis samples.

\improvedMCTS supports belief-dependent rewards by accumulating conditional
beliefs across multiple visitations of the same history (i.e. same node in the
belief tree). The estimated weight of each conditional belief is the sample
frequency of the corresponding hypothesis. That is,
$\hat{\mathbb{P}}(\beta_{0:t}^{i,j}\mid H_t)\!\triangleq\!\hat{\omega}^{i,j}_t \!=\!
\frac{\sum_{i,j}\mathbf{1}_{\beta=\beta_{0:t}^{i,j}}}{N}$, where $N$ is the number of
hypothesis samples, $i,j\in[1,|\beta_{0:t}|]$, $|\beta_{0:t}|$ is the
theoretical number of hypotheses at time $t$ and $\mathbf{1}_{\square}$ denotes
the indicator function.

\textbf{Parametric.} Assuming a parametric representation for the conditional
beliefs, $b[X_t]_{\beta_{0:t}}^{i,j}$, the belief-dependent reward,
$\mathcal{R}_b(b_t,a_t)$, is evaluated using the estimated hybrid belief,
$\mathcal{R}_b(\hat{b}_t,a_t)$, where
$\hat{b}_t=b[X_t]_{\beta_{0:t}}\hat{b}[\beta_{0:t}]\equiv
b[X_t]_{\beta_{0:t}}\hat{\mathbb{P}}(\beta_{0:t}\mid H_t)$, and $b_t$ defined in
\eqref{eq: hybrid belief}. Applying the
hypothesis resampling approach as described in Section \ref{sec: state dept
reward}, the sample frequency of each hypothesis in $\hat{b}_t$ is unbiased, in
other words, in expectation it equals the theoretical weights. Moreover,
\begin{lemma} \label{lemma4 - consistency of b dept. reward}
    $\mathcal{R}_b(\hat{b}_t,a_t)$ converges in probability to
    $\mathcal{R}_b(b_t,a_t)$ for any continuous, real-valued function
    $\mathcal{R}_b$.

    \begin{proof}
        By the law of large numbers, $\hat{\omega}_t^{i,j}$ is consistent as
        $N\rightarrow\infty$ for all ${i,j}\in[1,|\beta_{0:t}|]$,
        \begin{equation}
            \hat{\omega}_t^{i,j} 
                =\sum_{k=1}^N \frac{\mathbf{1}_{ \beta^k=\beta_{0:t}^{i,j}}}{N}
                \rightarrow^p  \mathbb{P}(\beta^{i,j}_{0:t}\mid H_t) \!= \omega_t^{i,j},
        \end{equation}
        then, due to the continuous mapping theorem,
        \begin{align}
            \mathcal{R}_b(b[X_t]_{\beta_{0:t}}\hat{b}[\beta_{0:t}], a_t)
            \rightarrow^p\mathcal{R}_b(b[X_t]_{\beta_{0:t}}b[\beta_{0:t}], a_t), \notag
        \end{align}
        that is,
        $\mathcal{R}_b(\hat{b}_t,a_t)$ is a consistent estimator for $\mathcal{R}_b(b_t,a_t)$.
    \end{proof}
\end{lemma}

\textbf{Nonparametric.} In the nonparametric case, the reward value is estimated
based on state particles, which may correspond to conditional belief estimation
via particle filters, or POMDPs with reward functions that have no close-form
solution, and are thus approximated via Monte Carlo methods. Then, instead of
$\mathcal{R}_b(b_t,a_t)$, an estimator over the reward is used,
$\hat{\mathcal{R}}_b(\hat{b}[X_t]_{\beta_{0:t}}\hat{b}[\beta_{0:t}],a_t)$, where
both the belief and the reward functions are estimators. We denote
$\hat{b}[X_t]_{\beta_{0:t}}^{ k} = \sum\nolimits
_{i=1}^{n_x}\alpha_t^{i,k}\delta(X-X_t^{i,k})$, where $\alpha_t^{i,k}$ is the
weight of state particle $i$ generated from conditional belief $k$ and $n_x$ is
the number of particles used to approximate the conditional belief. To arrive at
consistency results for an arbitrary nonparametric reward estimator, we assume
that the reward estimator based on samples from the full theoretical belief is
consistent, i.e.,
$\hat{\mathcal{R}}_b(\hat{b}[X_t]_{\beta_{0:t}}b[\beta_{0:t}],a_t)\rightarrow^p
\mathcal{R}_b(b_t,a_t)$.

\begin{lemma} \label{lemma4}
    If $\hat{\mathcal{R}}_b(\hat{b}[X_t]_{\beta_{0:t}}b[\beta_{0:t}],a_t)\rightarrow^p \mathcal{R}_b(b_t,a_t)$, then
    $\hat{\mathcal{R}}_b(b[X_t]_{\beta_{0:t}}\hat{b}[\beta_{0:t}],a_t)
    \rightarrow^p \mathcal{R}_b(b_t,a_t)$.

    \begin{proof}
        The proof follows similar steps to lemma \ref{lemma4 - consistency of b dept. reward}.
    \end{proof}
\end{lemma}

\subsection{Value function}
When using the existing hypotheses pruning approximations, the estimated value
function converges to the wrong value even when some external source provides
the exact reward value. This is due to the way observations are generated. The
value function is defined as
\begin{equation}
V^{\pi}(b_t) = \int_z \mathbb{P}(z_{t+1:\tau} \mid H_t^-) \sum _{\tau =t}^{\mathcal{T}}\mathcal{R}( b_{\tau } ,\pi_{\tau }) dz , 
\end{equation}
and since there is usually no direct access to observations given history, first
state-samples are generated, then observations are sampled using the observation
model, that is, $\mathbb{P}(z_{t} \mid H_t^-) = \sum_\beta \int_X \mathbb{P}(z_t
\mid X_t,\beta_{0:t}) b^-(X_t,\beta_{0:t})$. Replacing $b^-$ with its pruned
counterpart, $\hat{b}^-$, results in a shifted distribution for both the belief
and the measurements, which impacts the value function estimation. Proof of this
claim is similar to that of lemma \ref{lemma1} and skipped here for conciseness.

Instead, \improvedMCTS generates observations by first receiving a hypothesis
from the belief at the current node, $\beta_{0:t}^j$. Conditioned on
$\beta_{0:t}^j$ and the history, \improvedMCTS samples a new plausible
hypothesis, $\beta^i_{t+1}$. Then, an observation is sampled based on the
posterior hypothesis. More formally, 
\begin{align}
    &\mathbb{E}_{z_{t+1:\tau}} [ \sum_{\tau=t+1}^\mathcal{T} \mathcal{R}_{\tau} ]\! = \!
    \mathbb{E}_{z_{t+1}} \left[ \mathcal{R}_{t+1} + \mathbb{E}_{z_{t+2:\tau}}\left[V^\pi_{t+2}\right]\right] \\
    &= \underbrace{\mathbb{E}_{\beta_{0:t}} \mathbb{E}_{\beta_{t+1}\mid \beta_{0:t}} \mathbb{E}_{z_{t+1}\mid \beta_{0:t+1}} 
    \left[ \mathcal{R}_{t+1}\right]}_{\triangleq \alpha_{t+1}}
    + \mathbb{E}\left[V^\pi_{t+2}\right]\!. \notag
\end{align}
We then define the estimator for the expected reward, $\hat{\alpha}_{t+1}$,
\begin{align}
     \hat{\mathbb{E}}_{\mathbb{Q}}\!\!\left[\frac{\mathbb{P}\left( \beta _{t+1}^{i} \mid \beta _{0:t}^{j} ,H_{t+1}^{-}\right)}{\mathbb{Q}\left( \beta _{t+1}^{i} \mid \beta _{0:t}^{j} ,H_{0}\right)} \lambda _{t}^{j} \hat{\mathbb{E}}_{z_{t+1} \mid \beta _{0:t+\!1} ,H_{t+\!1}^{-}}[\hat{\mathcal{R}}_{t+1}]\right]
\end{align}
\begin{lemma} \label{lemma5} Given an unbiased reward estimator,
    $\hat{\mathcal{R}}$, the value-function estimator used in \improvedMCTS is
    unbiased.
\end{lemma}
\begin{proof} \label{proof:lemma5} Applying similar steps from the proof of
    lemma \ref{lemma2} on $\hat{\alpha}_{t+1}$, leads to an unbiased value,
    $\alpha_{t+1}$. Continuing recursively on the value function yields the
    desired result. See \cite{Barenboim23ral_supplementary} for further details.
\end{proof}

\section{Negative information in ambiguous data association}\label{sec:negInfo}
\begin{table}[t]
    \begin{center} 
    \begin{scriptsize}
    \begin{tabular}{ |c|c|c|c|c| } 
        \hline
        $z^{\beta_{t,k}} \!=\!  \infty$ & $\beta_{t,k}\!>\! n_{z_t}$ & $(x^{r},l^{k})\! \in\! S.R.$ &
       $\mathbb{P}(z  \mid x,l)$ & $\mathbb{P}(\beta \mid x,l)$ \\
        \hline
        no & no & yes & $f(\cdot)$ & 1  \\
        \hline
        no & no & no & 0 & 0 \\
        \hline
        yes & yes & no & 1 & 1 \\ 
        \hline
        yes & yes & yes & 0 & 0 \\
        \hline
        no & yes & yes & $f(\cdot)$ & 0 \\
        \hline
        no & yes & no & 0 & 1 \\
        \hline
        yes & no & no & 1 & 0 \\ 
        \hline
        yes & no & yes & 0 & 1 \\ 
        \hline
    \end{tabular}
    \caption{\label{table:negative_information}Possible combinations when considering negative information.
    $z^{\beta_{t,k}}=\infty$ indicates no observation. Hypothesis element
    $\beta_{t,k}> n_{z_t}$ assumes that $x^r,l^k$ are out of the sensing range.
    $(x^{r},l^{k}) \in S.R.$ indicates that a specific realization is within the
    sensing range. $\mathbb{P}(z^{\beta_{t,k}}  \mid x^{r},l^{k})$ and
    $\mathbb{P}(\beta_{t,k} \mid x^{r},l^{k})$ indicate the likelihood of the
    models. Last, $f(\cdot)$ denotes the likelihood value of the observation
    sensor (e.g. Gaussian).}
\end{scriptsize}
\vspace{-15pt}
    \end{center}
\end{table}
Just like observations affect the hypotheses' weights, not receiving an expected
observation also affects the weights, commonly known as negative information. 
We build on previous work
\cite{Pathak18ijrr} which addresses hybrid Bayesian inference for ambiguous DA
and shows how the mathematical formulation naturally extends to include negative
information. We limit our discussion of negative information to the context of
landmark-based observations. We conjecture that this formulation can also be
adapted to arbitrary observations, but is out of the scope of this paper.

Negative information is based on not receiving an observation from a mapped
landmark. We denote $|L_t|\in \mathbb{N}$ as the number of mapped
landmarks at time instant $t$. This usually refers to the number of landmarks
that already exist in the agent state (but can be defined otherwise). We also
define observation as, $z_t=[z_t^1,...z_t^{|L_t|}]$. Note that there are $|L_t|$
observation elements in the observation, even though usually not all landmarks
can be observed at a single time step, as some might be out of the sensing range
due to limited field of view, occlusions, and so on. If at time $t$ only $n_{z_t} <
|L_t|$ landmarks are observed, we fill the rest of the observation array with
$z^k_t=\infty$, i.e., out of sensing range. Then, the observation array becomes
$z_t\!=\![z_t^1,...,z_t^{n_{z_t}},\infty,...,\infty]_{1\times |L_t|}$. The reason
for such uncommon inflation of the observation array will become clear shortly.

We define $\beta_{t}=[\beta_{t,1},....,\beta_{t,|L_t|}]$ as an array that
subscribes each landmark with some observation. For example, $\beta_{t,k}=1$
associates landmark $l^k$ with observation-element $z_t^1$ from $z_t$. Note that
by the definition of the observation array, $z_t^{\beta_{t,k}}=\infty$ for all
$\beta_{t,k} > n_{z_t}$, which does not correspond to any real observation.

Equipped with the definitions of $\beta_t$ and $z_t$, we now discuss the
adaptation of the observation and association models. We drop the
$\square^{i,j}$ notation to avoid notation overloading, the derivations below
are true for each hypothesis separately. In the landmark-based context, it is
common to further simplify the expression in \eqref{eq: recursive DA weight
update} by assuming conditional independency of an observation given the state
variables, to a product of observation models, $\mathbb{P}( z_{ t} \mid
X_{t},\beta_{t}) = \prod^{|L_t|}_{k=1}\mathbb{P}( z^{\beta_{t,k}}_{ t} \mid
x^r_t,l^k)$, where $x^r_t$ and $l^k$ are the current pose of the agent and
landmark $k$. For simplicity, we assume in this paper an ideal detection sensor,
in the sense that if a landmark is within range, the sensor will detect it.
Under this assumption, likelihood of obtaining an out-of-range observation
($z^{\beta_{t,k}}_t=\infty$), given that the landmark is within the sensing
range (denoted $S.R.$), is $\mathbb{P}(z^{\beta_{t,k}}_t=\infty\mid x^r_t,l^k\in
S.R.)=0$. However, obtaining an out-of-range observation given that the landmark
is indeed out of the sensing range, is $\mathbb{P}(z^{\beta_{t,k}}_t=\infty\mid
x^r_t,l^k \notin S.R.)=1$.

The association model, $\mathbb{P}(\beta_{t,k}\mid x_t^k, l^k)$, assigns a
probability to associate a landmark, $l^k$, with a specific observation index,
$\beta_{t,k}$. We define the likelihood of associating an out-of-sensing-range
landmark to an actual observation element (i.e. $\beta_{t,k} \leq n_{z_t}$), as
$\mathbb{P}(\beta_{t,k} \leq n_{z_t}\mid x_t^k, l^k\notin S.R.)=0$. Conversely,
associating a landmark that is within the sensing range, equals a nonzero value,
for simplicity defined here as a uniform distribution across all feasible
associations, $\frac{1}{n_{z_t}}$. We explicitly state all possible combinations
of state, association, and observation in table \ref{table:negative_information}.


\section{Experiments}\label{sec:experiments}
In this section we evaluate our approach, \improvedMCTS, considering multiple
hypotheses due to ambiguous DA. We compare our approach with the state of the
art algorithms, DA-BSP \cite{Pathak18ijrr} and PFT-DPW \cite{Sunberg18icaps}.
PFT-DPW is utilized here as a single hypothesis solver, as it does not
explicitly support multiple hypotheses beliefs. Its hypothesis is chosen based
on the hypotheses weights through sampling. While it is possible to modify
PFT-DPW to accommodate multiple hypotheses, we leave this for future research.
To make DA-BSP comparable to other algorithms, we adapted the algorithm to
support anytime planning by utilizing Monte-Carlo trajectory samples instead of
a full tree traversal. We also evaluated \vanillaMCTS as the ad-hoc baseline for
MCTS implementation with hybrid beliefs, see table \ref{table:naming}.

\begin{table}[t]
    \centering
    \small 
    \begin{tabularx}{\columnwidth}{cXX}
        \toprule
        \textbf{Algorithm} & \textbf{Hypotheses control} & \textbf{Estimator} \\
        \midrule
        \vanillaMCTS (\ref{sec:vanillaMCTS}) & pruning & biased  \\
        PFT-DPW \cite{Sunberg18icaps} & single hypo. & biased \\
        DA-BSP \cite{Pathak18ijrr} & pruning & biased \\ 
        \improvedMCTS \textbf{(ours)} & sampling & unbiased \\
        \bottomrule
    \end{tabularx}
    \caption[]{\label{table:naming}Algorithms examined in our experiments.}
    \vspace{-15pt}
\end{table}

\begin{table}[h]
    \centering
    \small 
    \begin{tabularx}{\columnwidth}{cXXX} 
        \toprule 
        & \textbf{Aliased matrix} & \textbf{Goal reaching} & \textbf{Kidnapped robot} \\
        \midrule
        HB-MCP (ours) & -585.2 & -716.8 & -323.7 \\
        vanilla-HB-MCTS & -909.6 & -939.4 & -349.5\\
        PFT-DPW & -961.8 & -1009.8 & -327.8\\
        DA-BSP & -979.5 & -931.5 & -330.4\\
        \bottomrule
    \end{tabularx}
    \caption{Comparison of algorithm performances on different scenarios. Results are based on a simulation study with 100 trials per scenario and algorithm.}
    \label{table:algorithm_performance}
\end{table}

In all cases, the experiments were done using GTSAM library \cite{Dellaert12gt}
with a python wrapper as an inference engine for each of the hypotheses. Most
current state-of-the-art online tree search planners rely on particle filters as
an inference mechanism. However, particle filters are limited in their ability
to support high-dimensional and correlated state spaces efficiently. Instead,
through GTSAM we modeled each conditional belief as nonlinear state space
model corrupted with multivariate Gaussian noise. We give more information
of the hyperparameter choice in the supplementary file
\cite{Barenboim23ral_supplementary}. In the experiments, we assumed a SLAM
setting, in which the map is not perfectly known, and the agent is only given a
noisy prior on the map and its own pose. Due to ambiguous data associations,
each measurement may be obtained from any of the surrounding landmarks within
the sensing range of the agent. As a result of the ambiguous data associations,
the full posterior belief becomes multi-modal, with discrete variables
representing different possible associations.
\begin{figure}[h]
	\begin{subfigure}{0.24\textwidth}
		\includegraphics[width=\textwidth]{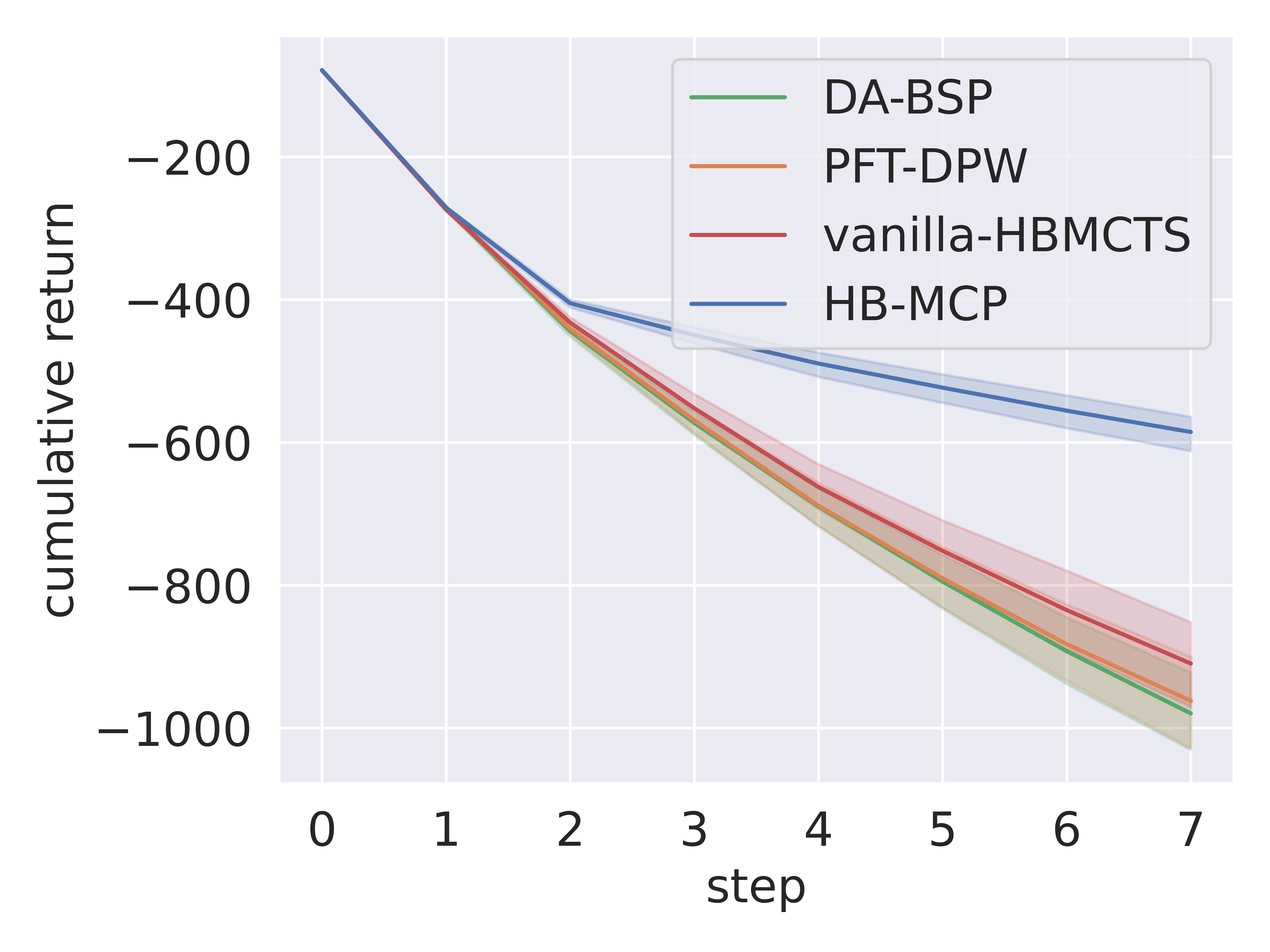}
		\caption{} 
	\end{subfigure}
	\begin{subfigure}{0.22\textwidth}
		\includegraphics[width=\textwidth]{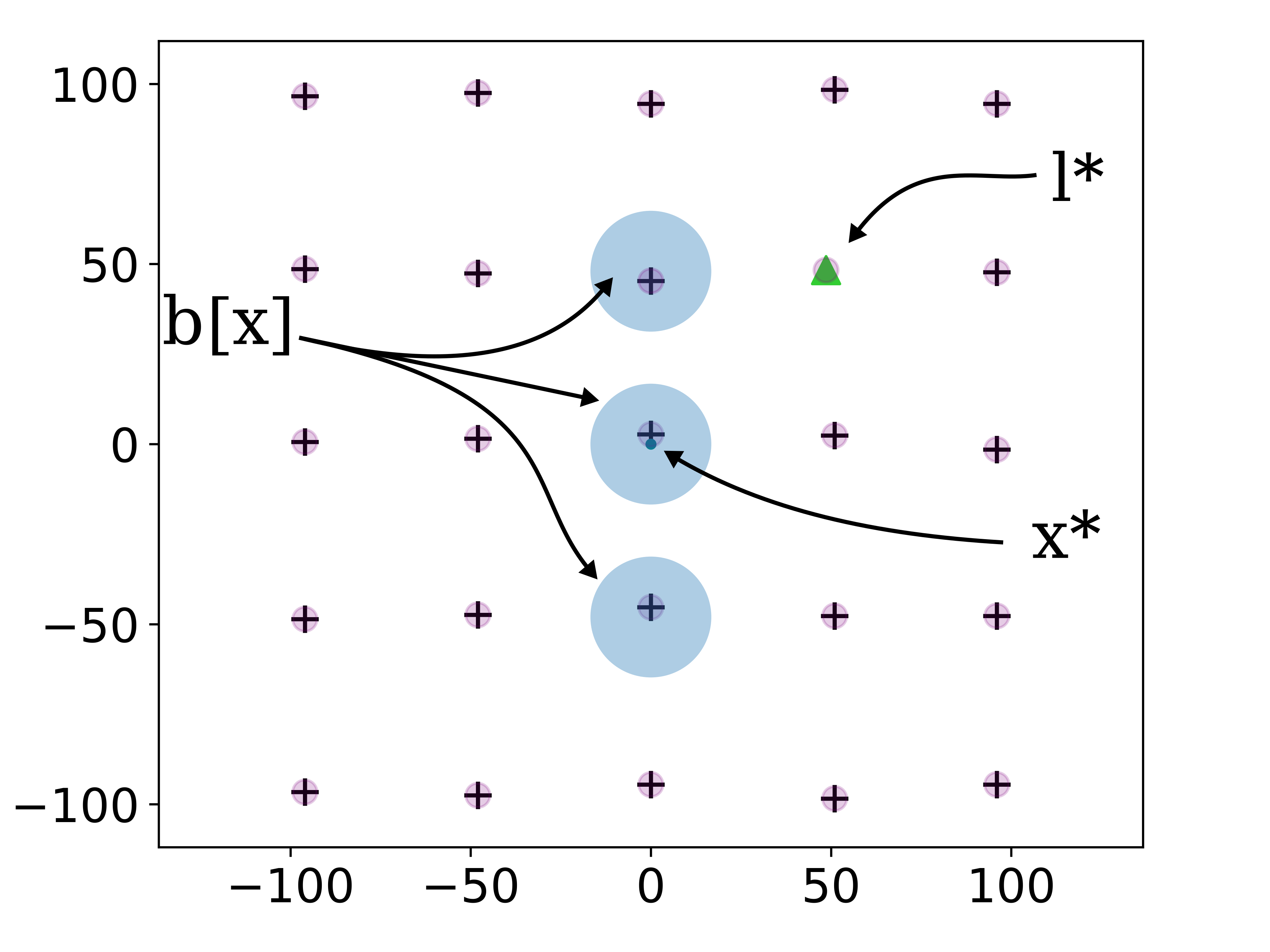}
		\caption{} 
	\end{subfigure}
	\caption{\textit{Aliased matrix.} The goal of the agent is to minimize the
	uncertainty of its pose and the location of all landmarks. (a) Mean and
	standard deviation of the cumulative reward, over 100 trials (higher is
	better). (b) Illustration of the initial belief of the agent. $x^*$ denotes
	the ground truth pose of the agent. $l^*$ denotes a unique landmark. The
	agent receives as a prior three hypotheses at different locations, drawn as
	blue ellipses.} \label{fig:ambigMatrix}
	\vspace{-10pt}
\end{figure} 

\textbf{Aliased matrix.} \label{sec:Ambiguous matrix} The first environment is a
highly aliased map, depicted in figure \ref{fig:ambigMatrix}(b). The task of the
agent is to reduce the uncertainty of its pose and all landmarks of the map,
measured by the (negative-) $\mathcal{A}$-optimality criteria. The
$\mathcal{A}$-optimality is the trace for the belief covariance matrix, commonly
used as uncertainty measure. The state of the agent is its trajectory and prior
landmarks. The agent is initially given three possible hypotheses for its pose,
and 24 aliased landmarks evenly scattered across the map and a unique landmark,
given as noisy prior to the agent. The unique landmark breaks the symmetry and
may be used by the agent to disambiguate hypotheses. The action space is defined
as a straight 4-directional open-loop actions, consisting of 12 intermediate
steps, each of 4[m]. Each planning session was limited to 40 seconds.

\begin{figure}[h]
	\begin{subfigure}{0.24\textwidth}
		\includegraphics[width=\textwidth]{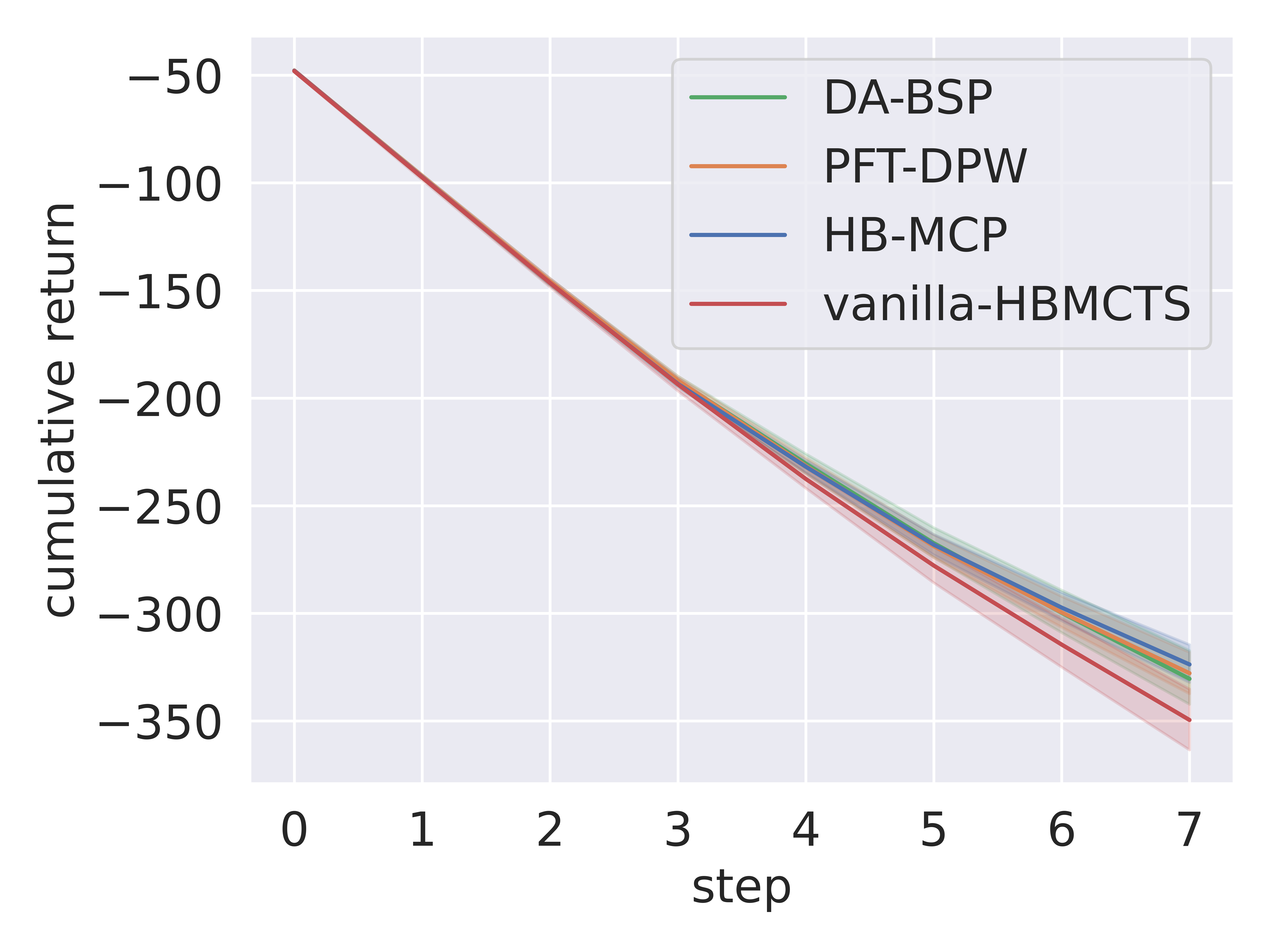}
		\caption{} 
	\end{subfigure}
	\begin{subfigure}{0.22\textwidth}
		\includegraphics[width=\textwidth]{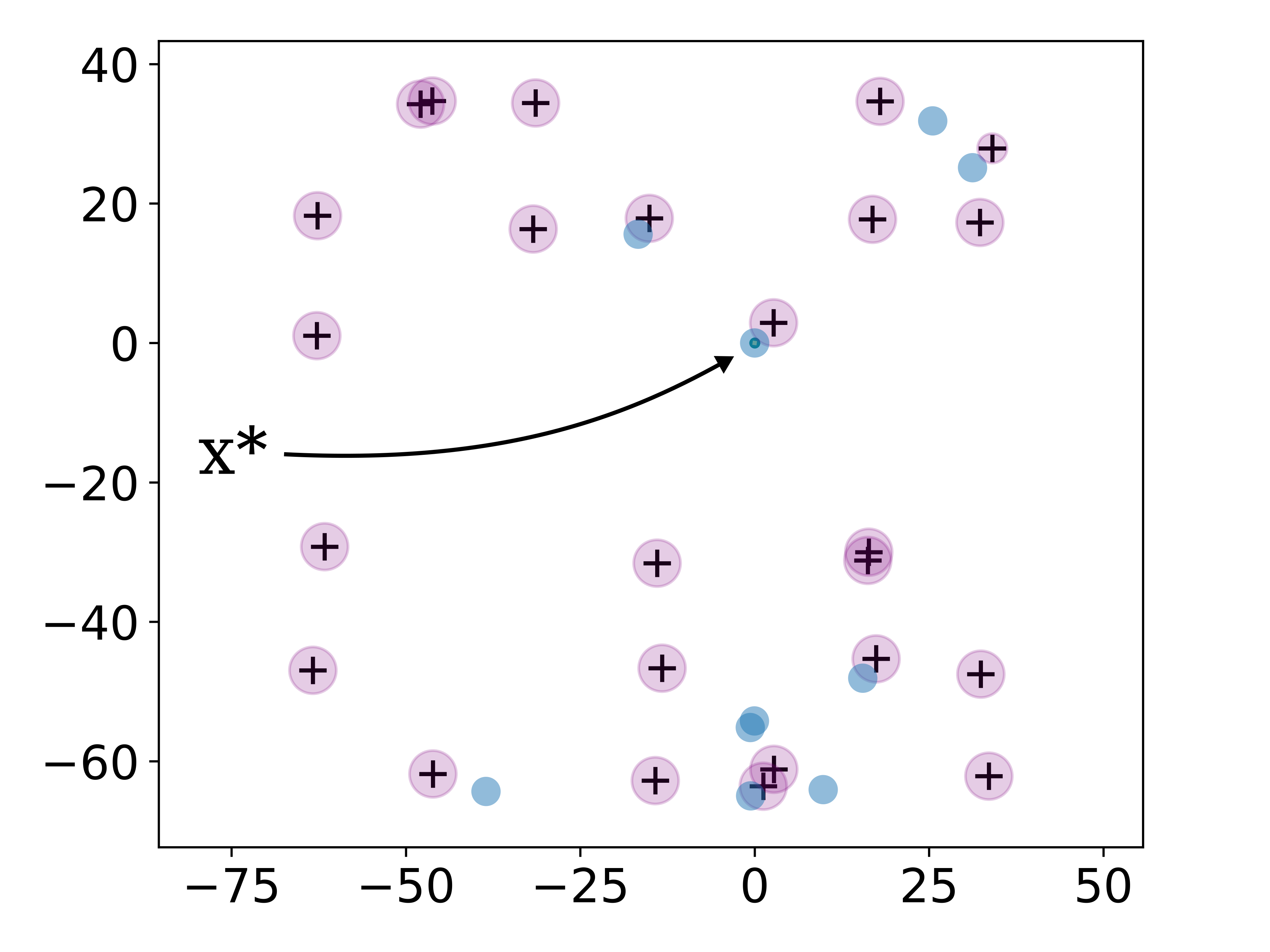}
		\caption{} 
	\end{subfigure}
	\caption{\textit{Kidnapped robot.} The goal of the agent is to minimize the uncertainty of its pose. (a) Mean and standard deviation of the cumulative reward, over 100 trials. (b) Illustration of the initial belief of the agent, blue circles illustrate conditional beliefs, crosses denote landmarks.}
	\label{fig:Kidnapped}
\end{figure} 
\textbf{Kidnapped robot.} The goal of the agent is to minimize the uncertainty
about the agent's pose. The environment has 16 randomly scattered landmarks on a
$160m \times 160m$ grid, with added Gaussian noise given as prior. The prior
pose of the agent is three hypotheses randomly scattered within the grid
boundaries. The action space is defined similarly to aliased matrix environment.
The reward function is defined by the $\mathcal{A}$-optimality criteria on the
robot's pose. Each planning session was limited to 20 seconds.
\begin{figure}[h]
	\begin{subfigure}{0.24\textwidth}
		\includegraphics[width=\textwidth]{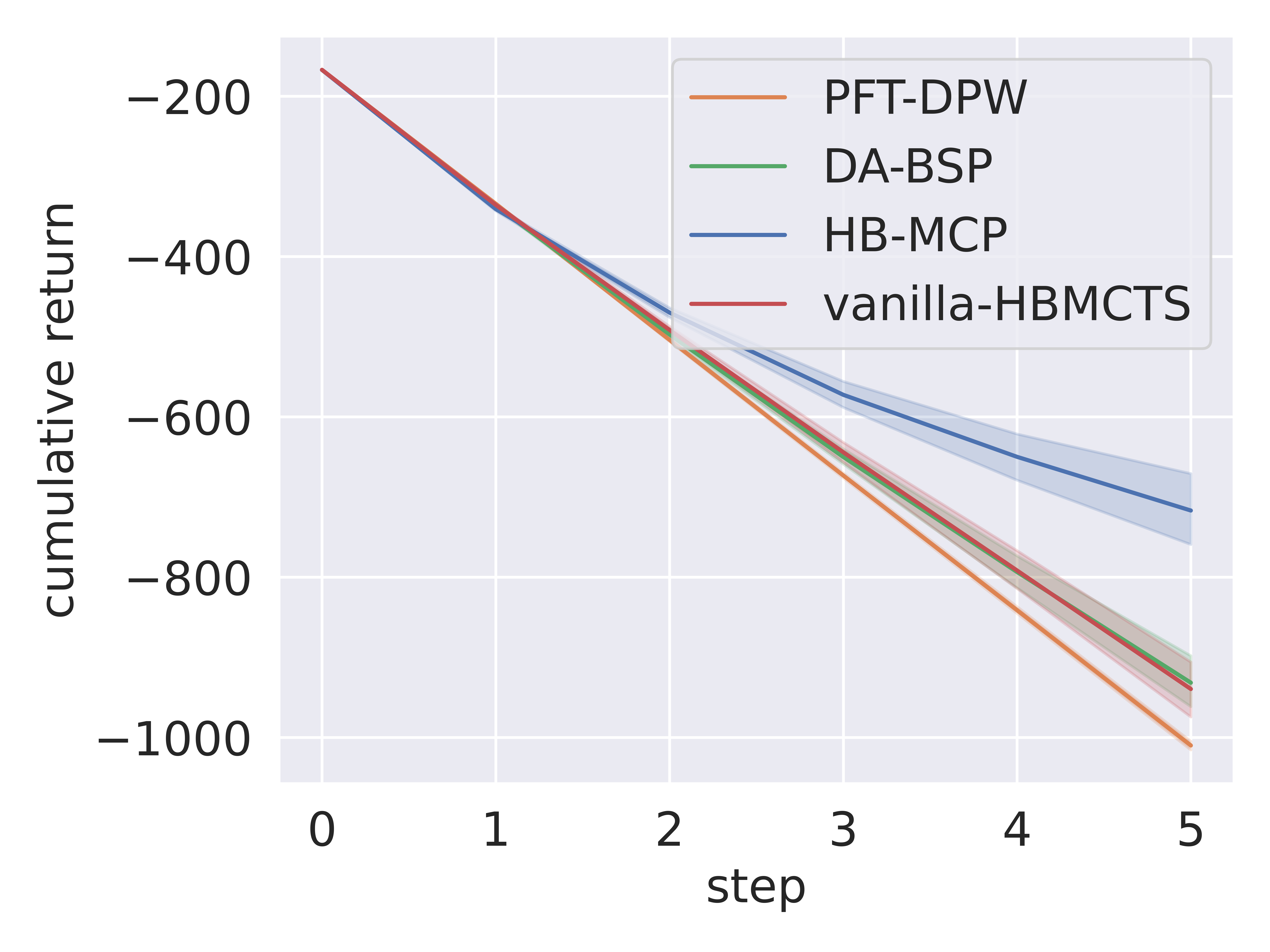}
		\caption{} 
	\end{subfigure}
	\begin{subfigure}{0.22\textwidth}
		\includegraphics[width=\textwidth]{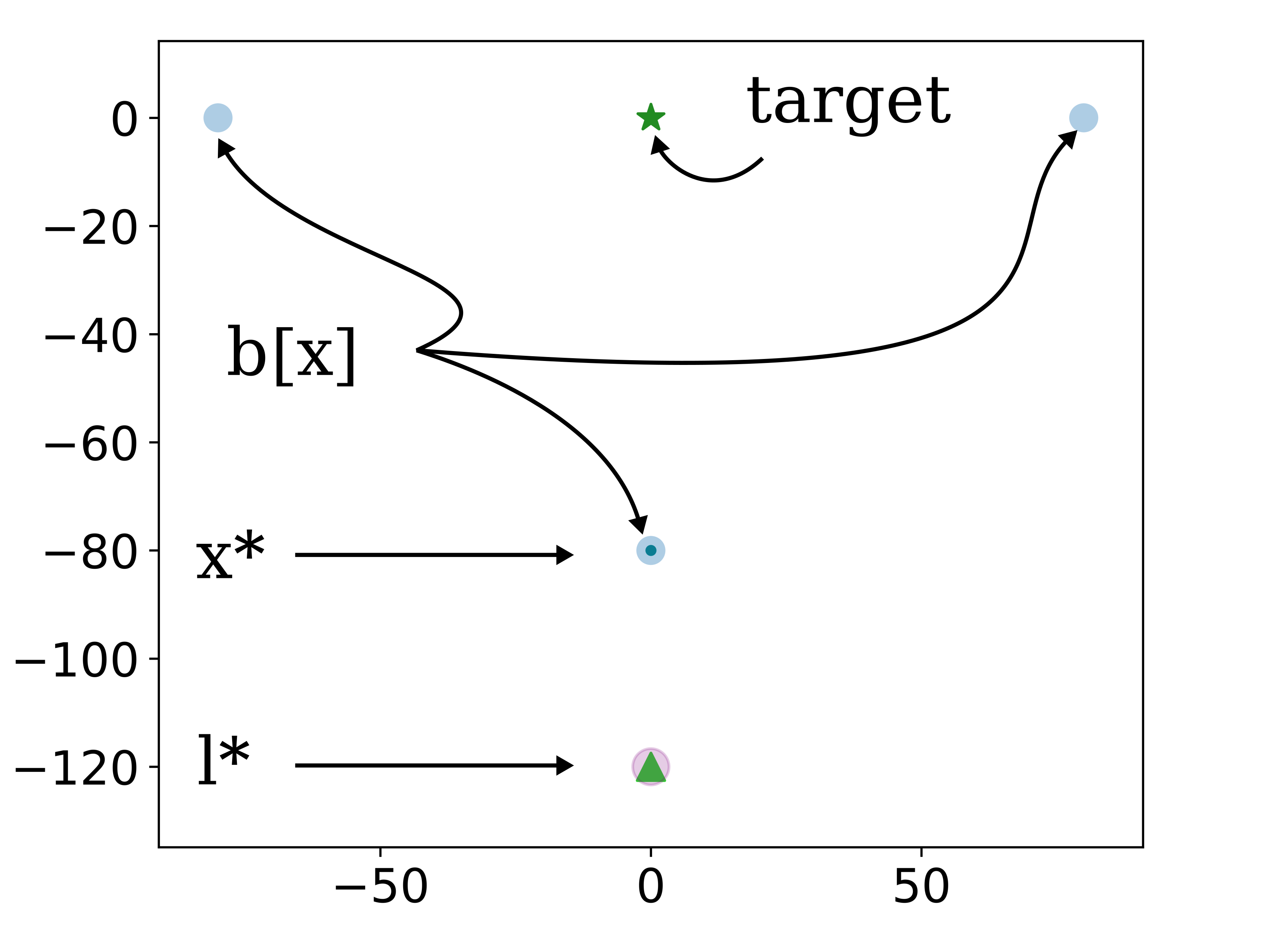}
		\caption{} 
	\end{subfigure}
	\caption{\textit{Goal reaching.} The goal of the agent is to reach the target location while
	minimizing uncertainty. (a) Mean and standard deviation of the cumulative
	reward, over 100 trials. (b) Illustration of the initial belief of the
	agent. $x^*$ denotes the ground truth pose of the agent. $l^*$ denotes a
	unique landmark. The agent receives as a prior three hypotheses at different
	locations. }
	\vspace{-15pt}\label{fig:goal reaching}
\end{figure} 

\textbf{Goal reaching.} The goal of the agent is to reach a predefined target
region. The agent prior belief is given as three hypotheses, located at
different directions with respect to the target. To ensure that the right
hypothesis gets to the target, the agent must first disambiguate some of the
hypotheses (using the unique landmark shown in figure \ref{fig:goal reaching}),
and only then attempt to reach the goal. The reward function is defined as the
negative sum of the Euclidean distance to goal and the $\mathcal{A}$-optimality
criteria. Each planning session was limited to 20 seconds.

\improvedMCTS received the highest expected cumulative reward in both the
ambiguous matrix and goal reaching scenarios. Note how in the ambiguous matrix
scenario, \improvedMCTS achieves significant improvement in cumulative reward
from step number 2. The reason for that is the agent's ability to spot the
unique landmark, which is two open-loop steps away when $t=0$, see figure
\ref{fig:ambigMatrix}(a). Due to restricted planning time, \vanillaMCTS and
\daBSP fail to identify and utilize the reduction in uncertainty via
disambiguation using the unique landmark. In all cases a single-hypothesis \PFT
is unaware of the multi-modality of the problem, and has no incentive to
prioritize the unique landmark over any other (ambiguous) landmark. In case of
\PFT, this statement is true for all the experiments.

In the kidnapped robot scenario the algorithms performed almost equally well,
with slight superiority to \improvedMCTS. Although \PFT is mathematically
inaccurate due to the choice of merely a single hypothesis, it enjoys higher
inference and planning efficiency which might translate is in some cases to good
performance. Although the kidnapped robot reward punishes for high uncertainty,
the random scatter of landmarks and poses did not lead to any strong preference
of a single policy for disambiguation, which can be clearly seen from the
cumulative reward of all algorithms in figure \ref{fig:Kidnapped} (a). Clearly,
depending on the scenario, even a heuristic, single-hypothesis solver might lead
to good performance. For more details, please refer to
\cite{Barenboim23ral_supplementary}.

\section{CONCLUSIONS}
In this work, we introduced \improvedMCTS, a novel algorithm to handle the
significant increase in computational effort of planning with hybrid
beliefs. We showed that current state-of-the-art algorithms rely on an
approximation, namely hypotheses pruning, that leads to a biased and
inconsistent reward and value function estimate. We proposed and analyzed a
different approach, namely \improvedMCTS, which utilizes sequential importance
resampling to converge to the correct value. Additionally, instead of building
symmetric hypotheses trees, \improvedMCTS focuses computations on the promising
branches corresponding to the UCB bonus. We demonstrated how \improvedMCTS could
be used for planning in ambiguous scenarios and derived a simple extension to
Bayesian inference to handle negative information naturally. Last, we
demonstrated our approach in a simulated environment. In our experiments,
\improvedMCTS outperformed the current state-of-the-art hybrid belief space
planning algorithms.

\addtolength{\textheight}{-12cm}   






\section*{ACKNOWLEDGMENT}
The authors thank Andrey Zhitnikov for helpful discussions regarding negative information.


\bibliographystyle{IEEEtran}
\bibliography{refs}

\end{document}


\maketitle

This document provides supplementary material to  \cite{Barenboim23ral_submitted}. Therefore,
it should not be considered a self-contained document, but instead regarded as an appendix of \cite{Barenboim23ral_submitted}. 
Throughout this report, all notations and definitions are with compliance to the ones presented in \cite{Barenboim23ral_submitted}. 

\section{Theoretical analysis}
\begin{lemma} \label{lemma2}
    HB-MCP state-dependent reward estimator,
    $\hat{\mathcal{R}}_X\triangleq \frac{1}{N}\sum_{i,j=1}^N \lambda_t^{i,j}
    \frac{1}{n_X}\sum_{k=1}^{n_X} r(X_t^{i,j,k},a_t)$, is unbiased.
    
    \begin{proof} If states are sampled i.i.d. for each hypothesis, then the
    expected value of the reward estimator, $\hat{\mathcal{R}}_X$, is,
\begin{align*}
    &\mathbb{E}[\hat{\mathcal{R}}] = \int \mathbb{Q}(\hat{\mathcal{R}}_X\mid H_t) \hat{\mathcal{R}}_X d\hat{\mathcal{R}}_X \\
    &= \int \int \int \mathbb{Q}(\hat{\mathcal{R}}_X,b, x_{1:n}\mid H_t) \hat{\mathcal{R}}_X dx_{1:n} db d\hat{\mathcal{R}}_X \\
    &= \int \int \int \mathbb{Q}(\hat{\mathcal{R}}_X\mid x_{1:n}) \mathbb{Q}(b, x_{1:n}\mid H_t) \hat{\mathcal{R}}_X dx_{1:n} db d\hat{\mathcal{R}}_X \\
    &= \int \int \int \mathbb{Q}(\hat{\mathcal{R}}_X\mid x_{1:n}) \mathbb{Q}(x_{1:n}\mid b, H_t) \mathbb{Q}(b \mid H_t) \hat{\mathcal{R}}_X dx_{1:n} db d\hat{\mathcal{R}}_X \\
    &= \int \int \mathbb{Q}(\hat{\mathcal{R}}_X\mid x_{1:n}) \mathbb{Q}(x_{1:n}\mid b_t, H_t) \hat{\mathcal{R}}_X dx_{1:n} d\hat{\mathcal{R}}_X \\
    &= \int \int \mathbb{Q}(\hat{\mathcal{R}}_X\mid x_{1:n}) \left[\sum_{i,j} \mathbb{Q}(x_{1:n}\mid b_t, \beta_{0:t}^{i,j}, H_t) \mathbb{Q}(\beta_{0:t}^{i,j}\mid b_t, H_t)\right]\hat{\mathcal{R}}_X dx_{1:n} d\hat{\mathcal{R}}_X 
\end{align*}
\begin{align*}
    &= \int \int \mathbb{Q}(\hat{\mathcal{R}}_X\mid x_{1:n}) \left[\sum_{i,j} \mathbb{Q}(x_{1:n}\mid b_t, \beta_{0:t}^{i,j}, H_t) \mathbb{Q}(\beta_{0:t}^{i,j}\mid H_t)\right]\hat{\mathcal{R}}_X dx_{1:n} d\hat{\mathcal{R}}_X \\
    &= \int \hat{\mathcal{R}}_X(x_{1:n}) \left[\sum_{i,j} \mathbb{Q}(x_{1:n}\mid b_t, \beta_{0:t}^{i,j}, H_t) \mathbb{Q}(\beta_{0:t}^{i,j}\mid H_t)\right] dx_{1:n} \\
    &= \sum_{i,j} \mathbb{Q}(\beta_{0:t}^{i,j}\mid H_t) \int \mathbb{Q}(x_{1:n}\mid b_t, \beta_{0:t}^{i,j}, H_t)  \hat{\mathcal{R}}_X(x_{1:n}) dx_{1:n} \\
    &= \sum_{i,j} \mathbb{Q}(\beta_{0:t}^{i,j}\mid H_t) \int \mathbb{Q}(x_{1:n}\mid \beta_{0:t}^{i,j}, H_t)  \hat{\mathcal{R}}_X(x_{1:n}) dx_{1:n} \\
    &= \mathbb{E}_{\mathbb{Q}} \mathbb{E}_{b[X_t]_{\beta_{0:t}}} \hat{\mathcal{R}}_X(x_{1:n}) =
    \mathbb{E}\left[\frac{1}{N}\sum_{i,j=1}^N \lambda_t^{i,j} \frac{1}{n_X}\sum_{k=1}^{n_X} r(X_t^{i,j,k},a_t)\right] \\
    &=\mathbb{E}_{\mathbb{Q}}\left[\frac{1}{N}\sum_{i,j=1}^N \lambda_t^{i,j} \mathbb{E}_{b[X_t]_{\beta_{0:t}}^{i,j}}\left[\frac{1}{n_X}\sum_{k=1}^{n_X} r(X_t^{i,j,k},a_t)\right]\right] \\
    &= \frac{1}{N}\sum_{i,j=1}^N \mathbb{E}_{\mathbb{Q}} \left[ \frac{\mathbb{P}}{\mathbb{Q}} \frac{1}{n_X}\sum_{k=1}^{n_X}\mathbb{E}_{b[X_t]_{\beta_{0:t}}}\left[ r(X_t^{i,j,k},a_t)\right] \right]\notag \\
    &= \mathbb{E}_{\mathbb{P}} \left[\mathbb{E}_{b[X_t]_{\beta_{0:t}}} r(X_t,a_t) \right]  \triangleq \mathcal{R}_X
\end{align*}
where $\mathbb{P}\!=\!\mathbb{P}(\beta_{0:t}\mid H_t)$,
$\mathbb{Q}\!=\!\mathbb{Q}(\beta_{0:t}\mid H_t)$, and $N$ and $n_X$ denote the number
of samples from $\mathbb{Q}$ and $b[X_t]^{i,j}_{\beta_{0:t}}$ respectively.
\end{proof}
\end{lemma}

\begin{lemma} \label{lemma5} Given an unbiased reward estimator,
    $\hat{\mathcal{R}}$, the value-function estimator used in HB-MCP is
    unbiased.
\end{lemma}
\begin{proof} \label{proof:lemma5}
    First, note that the value function of time step $t+1$ can be written as, 
    \begin{align}
        &\mathbb{E}_{z_{t+1:\tau}} [ \sum_{\tau=t+1}^\mathcal{T} \mathcal{R}_{\tau} ]\! = \!
        \mathbb{E}_{z_{t+1}} \left[ \mathcal{R}_{t+1} + \mathbb{E}_{z_{t+2:\tau}}\left[V^\pi_{t+2}\right]\right] \\
        &= \underbrace{\mathbb{E}_{\beta_{0:t}} \mathbb{E}_{\beta_{t+1}\mid \beta_{0:t}} \mathbb{E}_{z_{t+1}\mid \beta_{0:t+1}} 
        \left[ \mathcal{R}_{t+1}\right]}_{\triangleq \alpha_{t+1}}
        + \mathbb{E}\left[V^\pi_{t+2}\right]\!. \notag
    \end{align}
    and its corresponding estimator,
    \begin{align}
        \hat{\alpha}_{t+1}\triangleq\hat{\mathbb{E}}_{\mathbb{Q}}\!\!\left[\frac{\mathbb{P}\left( \beta _{t+1}^{i} \mid \beta _{0:t}^{j} ,H_{t+1}^{-}\right)}{\mathbb{Q}\left( \beta _{t+1}^{i} \mid \beta _{0:t}^{j} ,H_{0}\right)} \lambda _{t}^{j} \hat{\mathbb{E}}_{z_{t+1} \mid \beta _{0:t+\!1} ,H_{t+\!1}^{-}}[\hat{\mathcal{R}}_{t+1}]\right].
    \end{align}
    Then,
    \begin{gather*}
        \mathbb{E}[ \hat{\alpha}_{t+1}] =\mathbb{E}\left[\hat{\mathbb{E}}_{\beta _{0:t+1}^{i,j} \mid H_{t+1}^{-} \sim \mathbb{Q}}\left[\frac{\mathbb{P}\left( \beta _{t+1}^{i} \mid \beta _{0:t}^{j} ,H_{t+1}^{-}\right)}{\mathbb{Q}\left( \beta _{t+1}^{i} \mid \beta _{0:t}^{j} ,H_{0}\right)} \lambda _{t}^{j} \cdot \hat{\mathbb{E}}_{z_{t+1} \mid \beta _{0:t+1} ,H_{t+1}^{-}}[\hat{\mathcal{R}}_{t+1}]\right]\right]\\
        =\mathbb{E}\left[\frac{1}{N}\sum _{i=1}^{N}\frac{\mathbb{P}\left( \beta _{t+1}^{i} \mid \beta _{0:t}^{j} ,H_{t+1}^{-}\right)}{\mathbb{Q}\left( \beta _{t+1}^{i} \mid \beta _{0:t}^{j} ,H_{0}\right)}\frac{\mathbb{P}\left( \beta _{0:t}^{j} \mid H_{t}\right)}{\mathbb{Q}\left( \beta _{0:t}^{j} \mid H_{0}\right)} \cdot \frac{1}{n_{z}}\sum _{k=1}^{n_{z}}\hat{\mathcal{R}}_{t+1}\right]\\
        =\frac{1}{N}\sum _{i=1}^{N}\mathbb{E}_{\mathbb{Q}}\left[\frac{\mathbb{P}\left( \beta _{0:t+1}^{i,j} \mid H_{t+1}^{-}\right)}{\mathbb{Q}\left( \beta _{0:t+1}^{i,j} \mid H_{0}\right)}\frac{1}{n_{z}}\sum _{k=1}^{n_{z}}\mathbb{E}_{z}\mathbb{E}_{\mathcal{R}}[\hat{\mathcal{R}}_{t+1}]\right]\\
        =\mathbb{E}_{\beta _{0:t+1} \mid H_{t+1}^{-} \sim \mathbb{P}}[\mathbb{E}_{z_{t+1} \mid \beta _{0:t+1} ,H_{t+1}^{-}}[\mathcal{R}_{t+1}]] =\mathbb{E}_{z_{t+1} \mid H_{t+1}^{-}}[\mathcal{R}_{t+1}]
    \end{gather*}
    Continuing recursively on the value function yields the desired result.
\end{proof}

\section{Implementation details - vanilla-HB-MCTS}
\begin{algorithm}[ht]
    {\scriptsize
        \caption{vanilla-HB-MCTS}
        \label{alg:Pruned-HBT}
        \textbf{Procedure}:\textsc{Simulate}($b,h,d$)
        \begin{algorithmic}[1] 
            \IF{d = 0}
            \RETURN 0
            \ENDIF
            \STATE $a \xleftarrow{} \underset{\bar{a}}{\arg \max} \ Q(b\bar{a}) +
            c\sqrt{\frac{log(N(b))}{N(b\bar{a})}}$
            \IF{$|C(ba)| \leq k_oN(ba)^{\alpha_o}$ }
            \STATE $b' \xleftarrow{}$ \textsc{PrunedPosterior}$(b, a)$
            \STATE $r \xleftarrow{}$ \textsc{Reward}$(b, a)$
            \STATE $C(ba)\cup \{(b',r)\}$
            \STATE $R \xleftarrow{} r +$\textsc{Rollout}$(b', d-1)$
            \ELSE{}
            \STATE $b',r \xleftarrow{}$ {Sample uniformly from $C(ba)$} 
            \STATE $R \xleftarrow{} r +$\textsc{Simulate}$(b', d-1)$
            \ENDIF
            \STATE $N(b) \xleftarrow{} N(b) + 1$
            \STATE $N(ba) \xleftarrow{} N(ba) + 1$
            \STATE $Q(ba) \xleftarrow{} Q(ba) + \frac{R-Q(ba)}{N(ba)}$
            \STATE \textbf{return} $R$
        \end{algorithmic}
    }
    \end{algorithm}
    
    \begin{algorithm}[ht]
            {\scriptsize
        \caption{PrunedPosterior}
        \label{alg:Pruned-Posterior}
        \textbf{Procedure}:\textsc{PrunedPosterior}($b,a$)
        {\\ \ \ \ \ \ \ \ \ \ \textit{ // $b \triangleq \{b_t^j, \omega_t^j\}_{j=1}^{M}$}}
        \begin{algorithmic}[1] 
            \STATE $z \leftarrow$  \textsc{SampleObservation}$(b, a)$
            \STATE $\{\omega_{t+1}^{i,j}\}_{i=1, j=1}^{L, M}  \xleftarrow{}\!$  \textsc{ComputeWeights}$(b, a, z)$ \textit{//eq.\eqref{eq: recursive weight update}}
            \STATE $\{\omega_{t+1}^{i,j}\}_{i=1, j=1}^{L^s(j), M} \xleftarrow{}$ \textsc{Prune}$(\{\omega_{t+1}^{i,j}\}_{i=1,j=1}^{L,M})$ 
            \STATE $\{\bar{\omega}_{t+1}^{i,j}\}_{i=1,j=1}^{L^s(j),M} \xleftarrow{}$  \textsc{Normalize}$(\{\omega_{t+1}^{i,j}\}_{i=1,j=1}^{L^s(j),M})$
            \FOR{$j\in [1,M]$}
                \FOR{$i\in [1,L^s(j)]$} 
                \STATE $b^{i,j}_{t+1} \xleftarrow{} \Psi(b^j_t,a,z,i)$ \textit{// eq.~\eqref{eq: conBeliefUp}}
                \STATE $b'${.append($\{b^{i,j}_{t+1}, \bar{\omega}^{i,j}_{t+1}\}$)}
                \ENDFOR
            \ENDFOR
            \RETURN $b'$
        \end{algorithmic}
    }
    \end{algorithm}
    
    Algorithms \ref{alg:Pruned-HBT} and \ref{alg:Pruned-Posterior} describe the main
    procedures of vanilla-HB-MCTS. Algorithm \ref{alg:Pruned-HBT} follows PFT-DPW
    \cite{Sunberg18icaps} closely. Line 3 in Algorithm \ref{alg:Pruned-HBT} performs
    action selection based on the UCT exploration bonus. In our experimental setting,
    we assumed discrete action space, and thus avoided action progressive widening,
    which can otherwise be replaced with Line 3. Line 4 performs observation
    progressive widening, which resamples previously seen observations. This step is
    required to avoid shallow trees due to a continuous observation space, see
    \cite{Sunberg18icaps} for further details. Algorithm \ref{alg:Pruned-Posterior}
    computes the pruned-posterior belief, given the multi-hypotheses posterior
    belief from the previous time-step and the selected action.

\section{Results}
This section is intended to provide more information about the experiments that
appear in the paper. Specifically, we provide the trajectories performed by
HB-MCP and attempt to interpret the results below. In table
\ref{table:MCTS_hyperparameters} we provide the hyperparameters used in our
experiments and in table \ref{table:algorithm_performance} we provide a numeric
values for the average cumulative reward of our experiments.

\begin{figure}[h]
    \centering
    \begin{subfigure}[b]{0.32\textwidth}
        \centering
        \includegraphics[width=\textwidth]{./figs/ambiguous_matrix.png}
        \caption{cumulative return}
        \label{fig:cumulative return matrix}
    \end{subfigure}
    \hfill
    \begin{subfigure}[b]{0.32\textwidth}
        \centering
        \includegraphics[width=\textwidth]{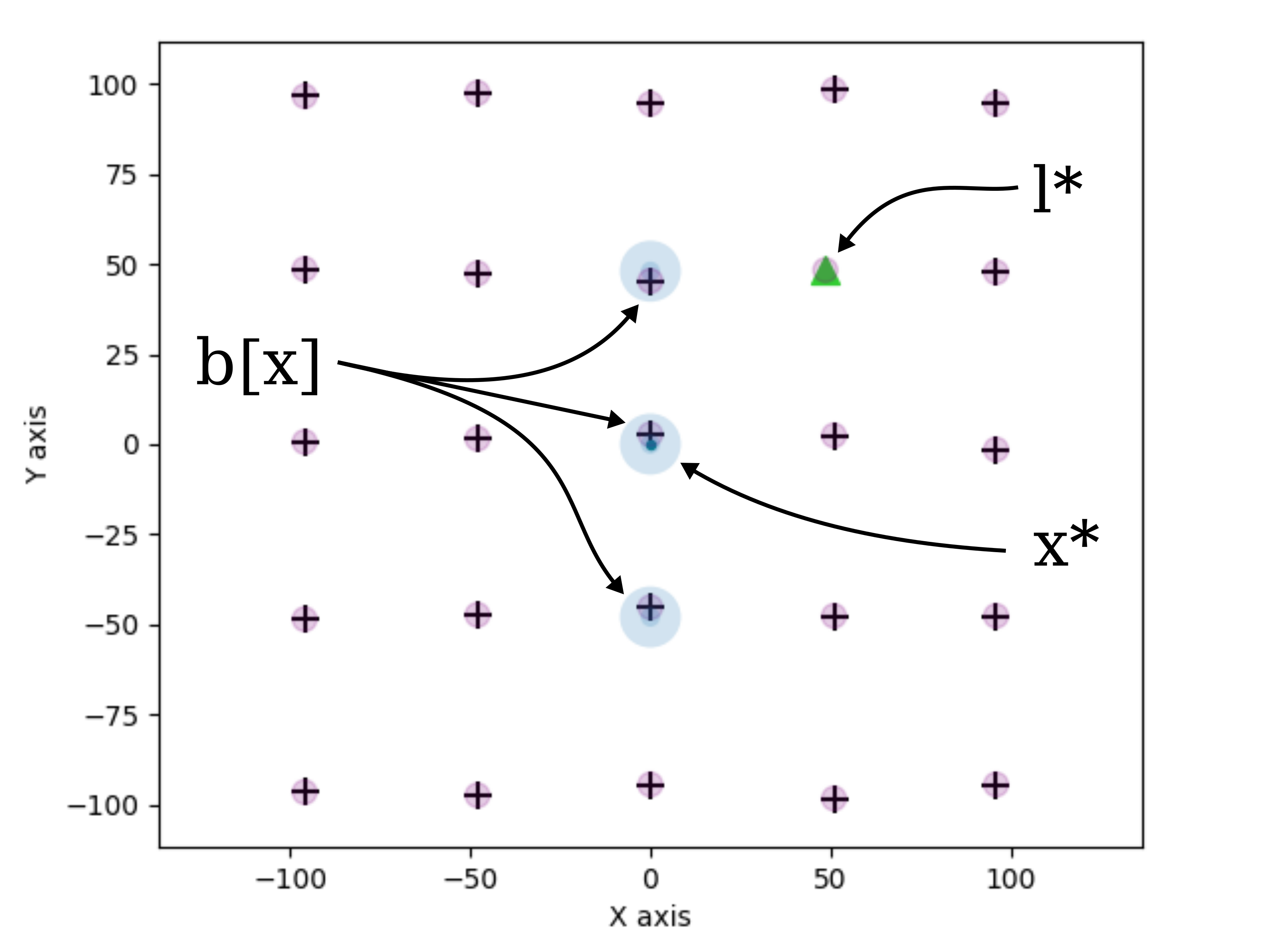}
        \caption{initial belief}
        \label{fig:initial belief matrix}
    \end{subfigure}
    \hfill
    \begin{subfigure}[b]{0.32\textwidth}
        \centering
        \includegraphics[width=\textwidth]{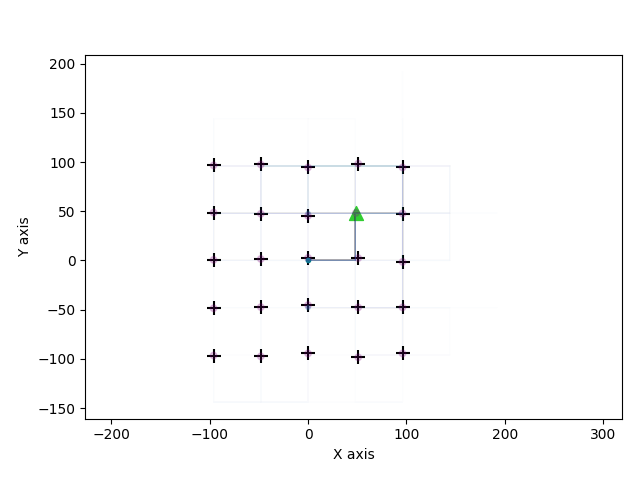}
        \caption{trajectories}
        \label{fig:trajectories matrix}
    \end{subfigure}
       \caption{The goal of the agent is to minimize the uncertainty of its pose and the location of all landmarks. (a) Mean and standard deviation of the cumulative reward, over 100 trials (higher is better). (b) Illustration of the initial belief of the agent. $x^*$ denotes the ground truth pose of the agent. $l^*$ denotes a unique landmark. The agent receives as a prior three hypotheses at different locations, drawn as blue ellipses. (c) Ground-truth trajectories are visualized in transparent color, illustrated on top of the initial belief, such that multiple similar trajectories appear in a moreopaque color.}
       \label{fig:three graphs}
\end{figure}

\textbf{Aliased matrix}. There are many ambiguous, evenly spaced landmarks
around the agent, along with its ambiguous initial pose, as shown in figure
\ref{fig:initial belief matrix}. The intuitive way to reduce the uncertainty of
the belief would be to first disprove wrong hypotheses, and then pass near as
many landmarks as possible, such that they would be within the sensing range.
The easiest way to disambiguate hypotheses would be to use the unique landmark
(see figure \ref{fig:initial belief matrix}). It is clearly shown in figure
\ref{fig:trajectories matrix} that the agent indeed prioritizes the unique
landmark before passing near landmarks. Note that the unique landmark would only
be visible (and thus provide observation) if the ground-truth position of the
landmark is within the sensing range of the ground-truth pose of the agent. It
can also be seen from figure \ref{fig:cumulative return matrix} that after two
macro-steps, which is the distance from the unique landmark, the descent in
cumulative reward becomes less steep, and significantly outperform other
algorithms.

\begin{figure}[h]
    \centering
    \begin{subfigure}[b]{0.32\textwidth}
        \centering
        \includegraphics[width=\textwidth]{./figs/goal_reaching.png}
        \caption{cumulative return}
        \label{fig:cumulative return goal}
    \end{subfigure}
    \hfill
    \begin{subfigure}[b]{0.32\textwidth}
        \centering
        \includegraphics[width=\textwidth]{./figs/env_goal_reachingII_arrows.png}
        \caption{initial belief}
        \label{fig:initial belief goal}
    \end{subfigure}
    \hfill
    \begin{subfigure}[b]{0.32\textwidth}
        \centering
        \includegraphics[width=\textwidth]{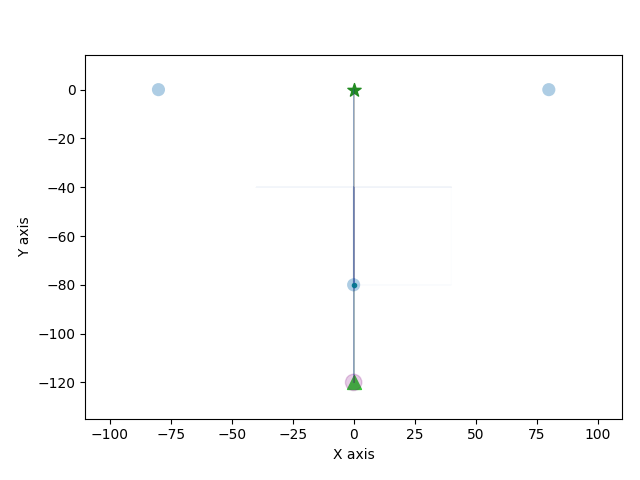}
        \caption{trajectories}
        \label{fig:trajectories goal}
    \end{subfigure}
       \caption{The goal of the agent is to reach the target location while minimizing uncertainty. (a) Mean and standard deviation of the cumulative reward, over 100 trials. (b) Illustration of the initial belief of the agent. $x^*$ denotes the ground truth pose of the agent. $l^*$ denotes a unique landmark. The agent receives as a prior three hypotheses at different locations. (c) Ground-truth trajectories are visualized in transparent color, illustrated on top of the initial belief, such that multiple similar trajectories appear in a moreopaque color.}
       \label{fig:three graphs}
\end{figure}
\textbf{Goal reaching}. As shown in \ref{fig:trajectories goal}, most of the
trajectories performed by the agent only walk through a simple straight line. Due
to the multi-modal hypotheses, the agent first prioritizes the unique landmark
(figure \ref{fig:initial belief goal}), which practically disambiguates wrong
hypotheses due to their large distance from the unique landmark. Then, the agent
chooses to reach the goal region to maximize the cumulative reward.

\begin{figure}[h]
    \centering
    \begin{subfigure}[b]{0.32\textwidth}
        \centering
        \includegraphics[width=\textwidth]{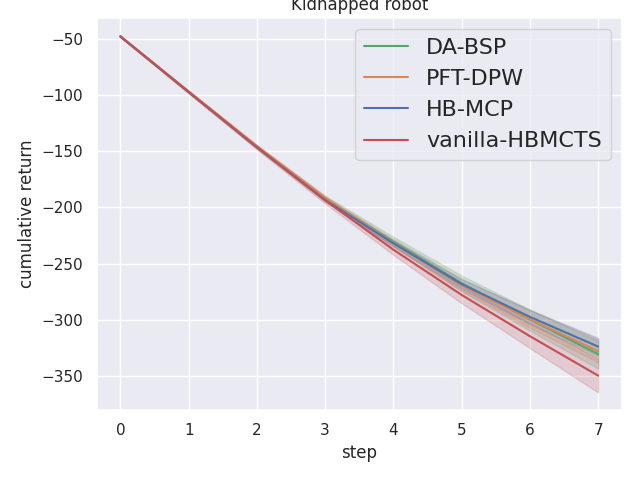}
        \caption{cumulative return}
        \label{fig:cumulative return kidnapped}
    \end{subfigure}
    \hfill
    \begin{subfigure}[b]{0.32\textwidth}
        \centering
        \includegraphics[width=\textwidth]{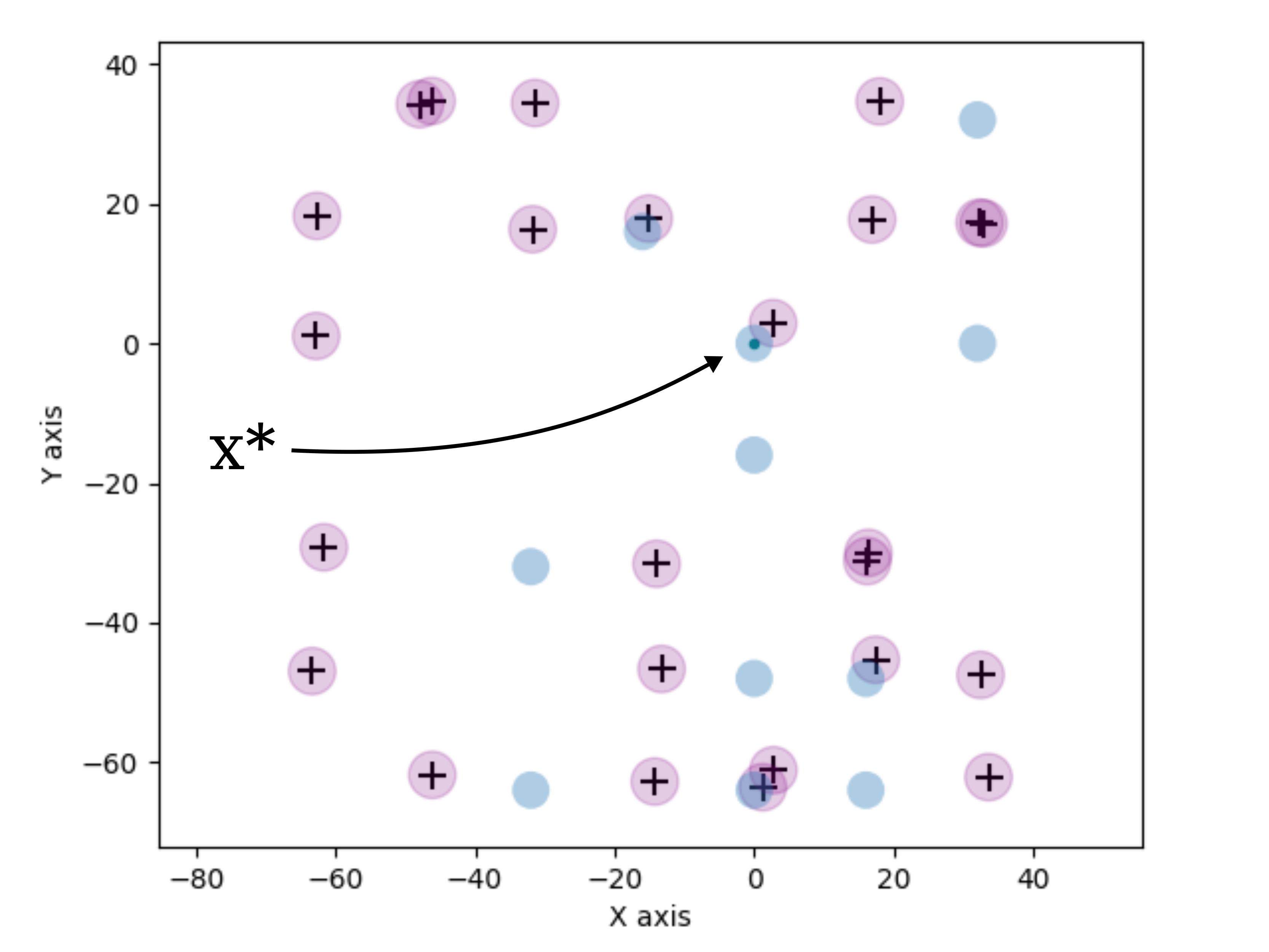}
        \caption{initial belief}
        \label{fig:initial belief kidnapped}
    \end{subfigure}
    \hfill
    \begin{subfigure}[b]{0.32\textwidth}
        \centering
        \includegraphics[width=\textwidth]{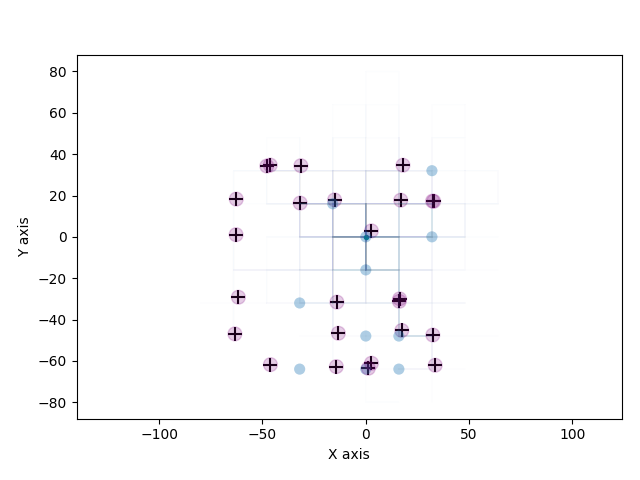}
        \caption{trajectories}
        \label{fig:trajectories kidnapped}
    \end{subfigure}
       \caption{The goal of the agent is to minimize the uncertainty of its pose. (a) Mean and standard deviation of the cumulative reward, over 100 trials. (b) Illustration of the initial belief of the agent, blue circles illustrate conditional beliefs, crosses denote landmarks. (c) Ground-truth trajectories are visualized in transparent color, illustrated on top of the initial belief, such that multiple similar trajectories appear in a moreopaque color.}
       \label{fig:three graphs}
\end{figure}

\textbf{Kidnapped robot}. The trajectories shown in figure \ref{fig:trajectories
kidnapped} do not show a strong preference to any direction. Note that the
environment is highly aliased, and there is no unique landmark where the agent
may reach to easily disprove wrong hypotheses. Similar results were obtained through
all solvers (figure \ref{fig:cumulative return kidnapped}). Although all landmarks
look alike, disambiguation may occur by utilizing the pattern of the scattered
landmarks. However, such disambiguation may require a long planning horizon which
was out of reach for our non-optimized planner. 

\begin{table}[h]
    \centering
    \begin{tabular}{|c|c|c|}
    \hline
    \textbf{Hyperparameter} & \textbf{Description} & \textbf{Default Value} \\
    \hline
    $c$ & UCB exploration constant & 40 \\
    \hline
    $N_{x}$ & Number of state particles per belief node & 200 \\
    \hline
    $T_{m}$ & Time limit per planning step (in seconds) & 20\footnote{} / 40\footnote{For Aliased
    matrix scenario} \\
    \hline
    $\mathcal{T}$ &     Lookahead horizon & 8 \\
    \hline
    $k_o$ & Observation double progressive widening multiplicative & 2.0 \\
    \hline
    $\alpha_o$ & Observation double progressive widening exponent & 0.014 \\
    \hline
    \end{tabular}
    \caption{Hyperparameters for HB-MCP (ours), vanilla-HB-MCTS and PFT-DPW
    algorithm. $^1$ indicates the planning time for Goal reaching and Kidnapped
    robot scenarios. $^2$ indicates the planning time for Aliased matrix
    scenario.}
    \label{table:MCTS_hyperparameters}
\end{table}

\begin{table}[h]
    \centering
    \begin{tabular}{|c|c|c|c|}
    \hline
     & \textbf{Aliased matrix} & \textbf{Goal reaching} &
    \textbf{Kidnapped robot} \\
    \hline
    HB-MCP (ours) & -585.2 & -716.8 & -323.7 \\
    \hline
    vanilla-HB-MCTS & -909.6 & -939.4 & -349.5\\
    \hline
    PFT-DPW & -961.8 & -1009.8 & -327.8\\
    \hline
    DA-BSP & -979.5 & -931.5 & -330.4\\
    \hline
    \end{tabular}
    \caption{Comparison of algorithm performances on different scenarios. Results are based on a simulation study with 100 trials per scenario and algorithm.}
    \label{table:algorithm_performance}
\end{table}

\bibliographystyle{IEEEtran}
\bibliography{refs}